\documentclass[sigconf]{acmart} 
\usepackage{fancyhdr}
\usepackage{subcaption}
\usepackage{tikz}
\usepackage{arydshln}
\usepackage{algorithm}
\usepackage{algpseudocode}
\algrenewcommand\textproc{\text}
\usepackage{booktabs}
\usepackage{color}
\usepackage{multirow}
\usepackage{enumitem}
\usepackage{balance}
\usepackage{makecell}
\usepackage{threeparttable}
\usepackage{amsmath,amsfonts,mathtools}
\usepackage{pifont}

\usepackage{mathrsfs}
\usepackage{float}
\usepackage{graphicx}
\usepackage{xcolor}
\usepackage{enumitem}
\usepackage{hyperref}
\usepackage{pifont}
\usepackage{marvosym}

\newcommand{\bluefont}[1]{ {\color{blue}{#1}}}
\newcommand{\redfont}[1]{{\textcolor{red}{#1}}}

\AtBeginDocument{%
  \providecommand\BibTeX{{%
    \normalfont B\kern-0.5em{\scshape i\kern-0.25em b}\kern-0.8em\TeX}}}

\copyrightyear{2025}
\acmYear{2025}
\setcopyright{acmlicensed}\acmConference[KDD '25]{Proceedings of the 31st ACM SIGKDD Conference on Knowledge Discovery and Data Mining V.1}{August 3--7, 2025}{Toronto, ON, Canada}
\acmBooktitle{Proceedings of the 31st ACM SIGKDD Conference on Knowledge Discovery and Data Mining V.1 (KDD '25), August 3--7, 2025, Toronto, ON, Canada}
\acmDOI{10.1145/xxxxxx.xxxx}
\acmISBN{978-1-4503-XXXX-X/18/06}

\newcommand{\M}{PatchSTG}

\begin{document}

\title{Efficient Large-Scale Traffic Forecasting with Transformers: A Spatial Data Management Perspective
}

\author{Yuchen Fang}
\affiliation{%
  \institution{
  University of Electronic Science and Technology of China}  
  \city{} 
  \country{}
  }
\email{fangyuchen@std.uestc.edu.cn}

\author{Yuxuan Liang}
\affiliation{
  \institution{
  The Hong Kong University of Science and Technology (Guangzhou)}  
  \city{}
  \country{}
  }
\email{yuxliang@outlook.com}

\author{Bo Hui}
\affiliation{%
\institution{Auburn University}
  \city{}
  \country{}
  }
\email{bzh0055@auburn.edu}

\author{Zezhi Shao}
\affiliation{%
\institution{Institute of Computing Technology,
Chinese Academy of Sciences}
  \city{}
  \country{}
  }
\email{shaozezhi@ict.ac.cn}

\author{Liwei Deng}
\affiliation{
\institution{University of Electronic Science and Technology of China}
\city{}
\country{}
}
\email{deng_liwei@std.uestc.edu.cn}

\author{Xu Liu}
\affiliation{
\institution{National University of Singapore}
\city{}
\country{}
}
\email{liuxu@comp.nus.edu.sg}

\author{Xinke Jiang}
\affiliation{
\institution{Peking University}
\city{}
\country{}
}
\email{thinkerjiang@foxmail.com}

\author{Kai Zheng}
\authornote{Corresponding author: Kai Zheng. He is with Yangtze Delta Region Institute (Quzhou), and School of Computer Science and Engineering, University of Electronic Science and Technology of China. The code is available at: https://github.com/LMissher/PatchSTG}
\affiliation{%
  \institution{
  University of Electronic Science and Technology of China}
  \city{}
  \country{}}
\email{zhengkai@uestc.edu.cn}

\renewcommand{\shortauthors}{Fang et al.}

\begin{abstract}
Road traffic forecasting is crucial in real-world intelligent transportation scenarios like traffic dispatching and path planning in city management and personal traveling. Spatio-temporal graph neural networks (STGNNs) stand out as the mainstream solution in this task. Nevertheless, the quadratic complexity of remarkable dynamic spatial modeling-based STGNNs has become the bottleneck over large-scale traffic data. From the spatial data management perspective, we present a novel Transformer framework called \M~to efficiently and dynamically model spatial dependencies for large-scale traffic forecasting with interpretability and fidelity. Specifically, we design a novel irregular spatial patching to reduce the number of points involved in the dynamic calculation of Transformer. The irregular spatial patching first utilizes the leaf K-dimensional tree (KDTree) to recursively partition irregularly distributed traffic points into leaf nodes with a small capacity, and then merges leaf nodes belonging to the same subtree into occupancy-equaled and non-overlapped patches through padding and backtracking. Based on the patched data, depth and breadth attention are used interchangeably in the encoder to dynamically learn local and global spatial knowledge from points in a patch and points with the same index of patches. Experimental results on four real world large-scale traffic datasets show that our \M~achieves train speed and memory utilization improvements up to $10\times$ and $4\times$ with the state-of-the-art performance.
\end{abstract}



\maketitle

\section{Introduction}
Road traffic data comprises multiple traffic time series collected from points where road sensors are deployed. Thus traffic time series are correlated in not only the temporal aspect but also spatial domain. Forecasting future road traffic through past data plays an essential role in many real world intelligent transportation applications. For instance, users on Map platforms can select the least time path in advance according to the predicted traffic~\cite{li2022mining,liu2022hega,liu2023impact,deng2024task}. Likewise, on city management platforms~\cite{deng2024million}, users can control the signal light to avoid congestion based on future traffic~\cite{fang2024cdgnet,lai2023large,sun2024crosslight}.

To accurately forecast future road traffic, countless algorithms have been proposed in past decades, ranging from statistical models~\cite{chandra2009predictions} to data-driven methods~\cite{jin2023spatio}. In the beginning, temporal modeling methods like the recurrent neural network (RNN) and autoregressive integrated moving average (ARIMA) are used to learn the temporal evolution of the single traffic time series in traffic data~\cite{kumar2015short,lv2018lc}, yet ignore the spatial transmission between multiple traffic time series. Subsequently, spatio-temporal graph neural networks~\cite{jiang2023uncertainty,wang2024comet,luo2024timeseries} are at the forefront of collaboratively capturing spatio-temporal dependencies.

However, the spatial correlation of traffic points is evolved over time and the fixed static graph in conventional spatio-temporal graph neural networks can not reflect various correlations in different time stages. Therefore, dynamic spatial modeling technology has been focused as the mainstream research line in traffic forecasting~\cite{guo2019attention}, which aims to reveal spatial correlations of each time slice respectively and dynamically propagate spatial information. Considering the quadratic complexity in most dynamic spatial modeling methods, traffic forecasting is only performed at zone scale, which is opposite to the realistic traffic forecasting needs in a city with thousands of traffic points~\cite{liu2024largest}.

\begin{figure}[t]
    \centering
        \begin{subfigure}{0.3\linewidth}
        \includegraphics[width=\linewidth]{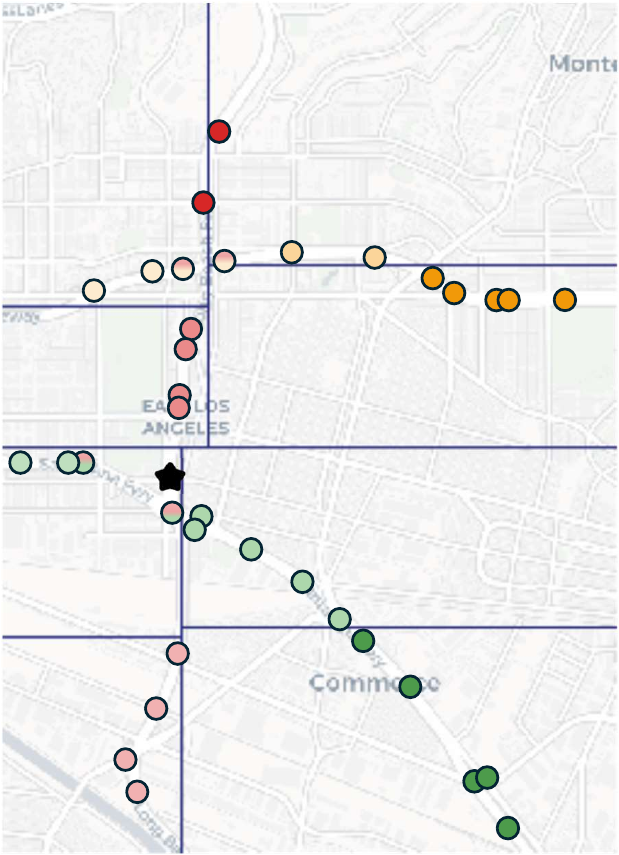}
        \subcaption{Partition}
        \label{intro:sp}
      \end{subfigure}%
      \hspace{1mm}
      \begin{subfigure}{0.3\linewidth}
        \includegraphics[width=\linewidth]{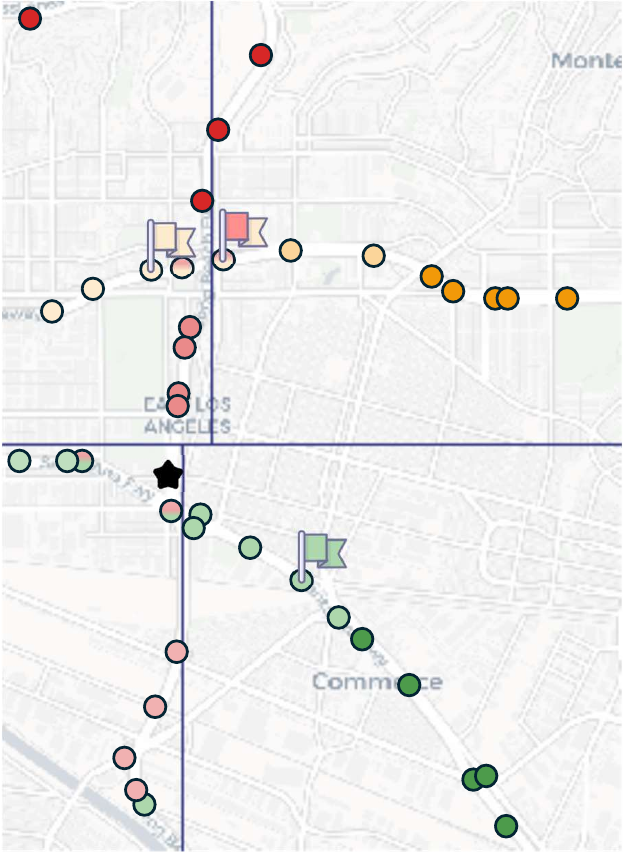}
        \subcaption{Patching}
        \label{intro:ss}
      \end{subfigure}
      \hspace{1mm}
      \begin{subfigure}{0.35\linewidth}
        \includegraphics[width=\linewidth]{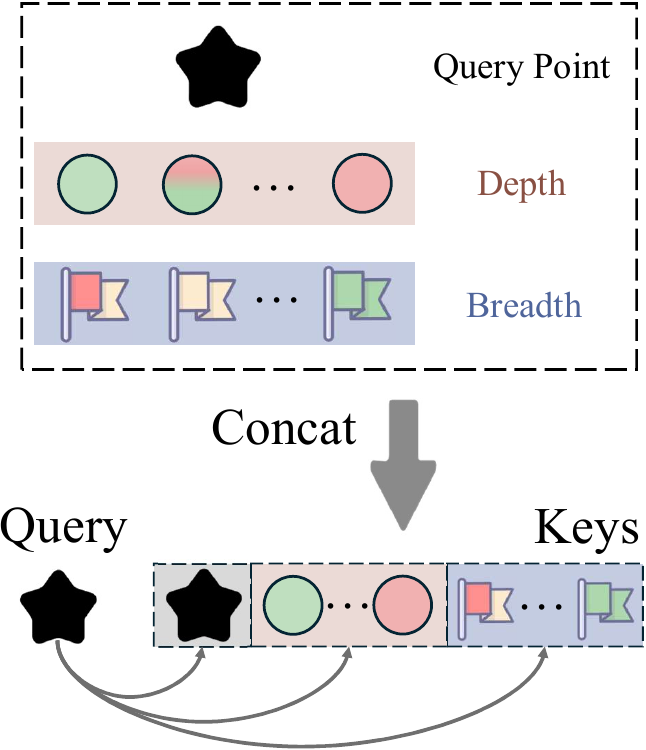}
        \subcaption{Attention}
        \label{intro:sa}
      \end{subfigure}
      \caption{Sketch of our \M. We first partition the equilibrium number of points into small regions. Then we merge small regions into patches after padding points into regions where the maximum number of points has not been reached. Finally, we perform the depth and breadth attention on points in the same patch and points in other patches with the same index to capture local and global spatial knowledge.}
      \label{intro:att}
\end{figure}

As shown in Table~\ref{intro:comp}, three widely used dynamic spatial modeling paradigms (dot-product, linear, and low-rank) are illustrated in the attention form, \emph{i.e.}, using query and key to dynamically calculate spatial correlations first and then propagating spatial information to the original value according to computed spatial correlations. Spatial correlations in dot-product-based methods such as D2STGNN~\cite{shao2022decoupled} and STAEformer~\cite{liu2023spatio} should be calculated on each pair of points thus leading to unacceptable quadratic complexity in large-scale traffic forecasting. Despite linear-based BigST~\cite{han2024bigst} and low-rank-based Airformer~\cite{liang2023airformer} being proposed to mitigate the high computation needs of dynamic spatial modeling in large-scale spatio-temporal forecasting, they still have some limitations. The drawback of linear-based~\cite{fang2022learning,han2024bigst} is the lack of interpretability because spatial correlations can not be explicitly shown. For low-rank-based~\cite{liang2021fine,fang2023stwave+}, the blemish is a lack of fidelity, \emph{i.e.}, crucial information cannot be guaranteed to remain in the reduced low-rank representations and thus results in a performance drop.

This work aims to reduce the computation needs of dynamic spatial modeling with high interpretability and without information loss in large-scale traffic forecasting by proposing an efficient Transformer framework called \M. The sketch of \M~ is illustrated in Figure~\ref{intro:att} and the core goal of \M~ is to reduce the number of points involved in dynamic calculations like vision Transformers~\cite{liu2021swin}. However, traffic points are irregularly distributed on roads, and thus simply splitting the same number of points into patches with equal size to reduce complexity in regular spatial data-based vision tasks is unsatisfactory in traffic forecasting. Inspired by the spatial data management algorithm KDTree (short for K-dimensional tree), we propose an irregular spatial patching method to split the same number of traffic points into patches with unequal size, which first uses the novel leaf KDTree to recursively partition all the traffic points into leaf nodes with the small capacity, and then merges leaf nodes belonging to the same subtree into occupancy-equaled and non-overlapped patches through padding unfull leaf nodes and backtracking to the root node of subtrees. Finally, based on the patched data, \M~first performs depth attention on points in a patch to learn local spatial knowledge and then conducts breadth attention on points in different patches but with the same index to efficiently aggregate multiple global knowledge. Notably, \M~ is interpretable and fidelity due to domain knowledge informed irregular spatial patching and non-compression dynamic spatial modeling.

\begin{table}[t]
    \aboverulesep=0ex
    \belowrulesep=0ex
    \caption{Four dynamic spatial modeling paradigms in traffic forecasting including previous dot-product, linear, low-rank, and our proposed patching. $N$, $R$, and $d$ denote the number of traffic points, the number of points after cluster or sampling, and the number of model dimensions. In general, $R\ll N$ and $d\ll N$. $Q$, $K$, $V$, $\bar{K}$, and $\bar{V}$ are query, key, value, reduced low-rank key, and reduced low-rank value in dynamic spatial modeling via attention form.}
    \centering    
    \resizebox{1.0\linewidth}{!}{
    \begin{tabular}{c|cccc}
    \toprule
        Paradigm & \multicolumn{1}{c}{Dot-Product} & \multicolumn{1}{c}{Linear} & \multicolumn{1}{c}{Low-Rank} & \multicolumn{1}{c}{Patching}\\
    \midrule
        Complexity & $O(N^2d)$ & $O(Nd^2)$ & $O(NRd)$ & $O(NRd)$\\
    \midrule
        Formulation & $(QK^T)V$ & $Q(K^TV)$ & $(Q\bar{K}^T)\bar{V}^T$ & $\mathop{\vert\vert}\limits^{R}_1(Q_{(i)}K^T_{(i)})V_{(i)}$\\
        \midrule
        \makecell[c]{Information\\Loss} & \ding{56} & \ding{56} & \ding{52} & \ding{56}\\
        \midrule
        Interpretability & \ding{52} & \ding{56} & \ding{52} & \ding{52}\\
        \midrule
        \makecell[c]{Domain\\Knowledge} & \ding{52} & \ding{56} & \ding{56} & \ding{52}\\
        \bottomrule
    \end{tabular}}
    \label{intro:comp}
\end{table}

In summary, we have made the following contributions:
\begin{itemize}[leftmargin=*]
    \item We present \M, a generic Transformer framework tailored for efficiently and effectively predicting large-scale traffic from the spatial data management perspective.
    \item  To the best of our knowledge, we are the first to bridge the gap between KDTree and patching technique to evenly partition irregularly distributed traffic points with interpretability.
    \item \M~ performs the depth and breadth attention on points in a patch and points with the same index to efficiently learn dynamic local and global spatial knowledge with fidelity.
    \item  Comprehensive experimental results on large-scale traffic datasets demonstrate that our \M~ achieves state-of-the-art performance and enjoys $4\times$ memory reduction and $10\times$ training speedup compared to dynamic spatial modeling baselines.
\end{itemize}

\section{Related Works}
\subsection{Traffic Forecasting}
Traffic forecasting has been a concern of research and industrial communities in past decades. Originally, the statistical-based vector autoregression (VAR)~\cite{chandra2009predictions} and autoregressive integrated moving average (ARIMA)~\cite{kumar2015short} are used to capture temporal dependencies. As deep learning is splendid in many tasks, recurrent neural network-based methods~\cite{lv2018lc} and temporal convolution network-based methods~\cite{guo2021traffic,hui2021trajectory} are proposed to improve traffic forecasting performance. To simultaneously extract spatio-temporal information, DCRNN~\cite{DBLP:conf/iclr/LiYS018} and STGCN~\cite{yu2018spa} constructed a fixed adjacency matrix in graph neural networks~\cite{yu2023learning,yu2024few,jiang2024ragraph} based on real-world distances to capture static spatial information for traffic forecasting. Subsequent techniques such as GWNET~\cite{wu2019graph}, AGCRN~\cite{bai2020adaptive}, MTGNN~\cite{wu2020connecting}, METRO~\cite{cui2021metro}, STG-NCDE~\cite{choi2022graph}, and Localised AGCRN~\cite{duan2023localised} have further improved forecasting accuracy through the data-driven fixed adjacency matrix. Nevertheless, most of them ignored the evolved spatial correlations of traffic. For calculating dynamic point-to-point spatial correlations, most approaches such as ASTGCN~\cite{guo2019attention}, GMAN~\cite{zheng2020gman}, ST-GRAT~\cite{park2020st}, DMSTGCN~\cite{han2021dynamic}, GMSDR~\cite{liu2022msdr}, and etc.~\cite{wang2023multi,jiang2023pdformer,liu2023spatio,li2024urban,yu2024dygprompt} applied the dot-product operation introduced by the attention mechanism on hidden representations with different periods. However, point-to-point dynamic models have brought the efficiency bottleneck into large-scale traffic forecasting. Despite efficient dynamic spatial modeling methods such as linear-based Lastjomer~\cite{fang2022learning}, BigST~\cite{han2024bigst} and low-rank-based HIEST~\cite{ma2023rethinking}, SSTBAN~\cite{guo2023self} have been proposed to reduce the complexity, explicit spatial correlations are failed to report in linear-based methods and performance is restricted in low-rank-based methods compared with perceptron-based STID~\cite{shao2022spatial} and SimST~\cite{liu2024reinventing} due to spatial reduction caused information loss. Therefore, we propose a novel efficient dynamic spatial modeling method \M, which is intepretable and fidelity.

\subsection{Efficient Spatial Transformers}
The quadratic complexity of dot-product attention has become the bottleneck in applying Transformers to learn spatial knowledge, thus efficient spatial Transformers have been researched in recent years. For regular spatial data such as images and videos, the same number of neighbored pixels can be simply merged into the same size patches such as ViT~\cite{dosovitskiy2020image} and SwinTransformer~\cite{liu2021swin} to decrease complexity by reducing points in the calculation. Different from regular spatial data, GraphTrans-ViT~\cite{he2023generalization} and PatchGT~\cite{gao2022patchgt} can only derive the overlapped and unbalanced patches through clustering algorithms. Luckily, STRN~\cite{liang2021fine}, FPT~\cite{park2022fast} and OctFormer~\cite{wang2023octformer} segmented images into patches with different sizes based on the semantic and distribution, which gave us the inspiration to use spatial data management algorithms for patching time-ordered irregular spatial data into balanced and non-overlapped patches.

\section{Preliminaries}
\textbf{Traffic Data.} Traffic data is made up of multiple correlated time series collected from points where road sensors are deployed. The recorded time series of the specific traffic point $n$ can be formulated as $x_n\in\mathbb{R}^{H}$, which contains traffic volume within $H$ time slices. Therefore, traffic data that comprises time series on $N$ points can be formed as a matrix $X\in\mathbb{R}^{H\times N}$, where $x_{n}^h$ denotes the traffic flow of point $n$ at time $h$.

\noindent\textbf{Traffic Forecasting.} In the traffic forecasting task, the common setting involves predicting future traffic features through historical values. Specifically, the goal of our paper is to forecast the future traffic flow for the next $F$ time slices according to information from the preceding $H$ time slices and the location of points.
\begin{equation}
    \hat{Y}=f_{\theta}(X,Lat,Lng)
\end{equation}
where $\hat{Y}\in\mathbb{R}^{F\times N}$ is the predicted traffic in the future, which will be used to compare with the ground truth $Y\in\mathbb{R}^{F\times N}$. $Lat\in\mathbb{R}^{N}$ and $Lng\in\mathbb{R}^{N}$ denote the real-world latitude and longitude of points. Moreover, the function $f_{\theta}(\cdot)$ indicates a data-driven forecasting model parameterized by $\theta$.

\begin{figure}[t]
  \centering
\includegraphics[width=\linewidth]{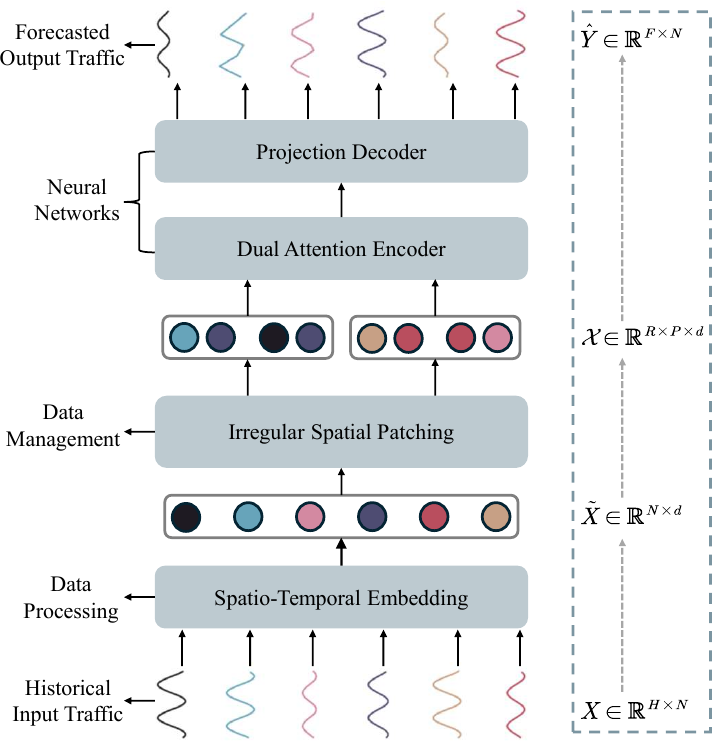}
  \caption{The workflow of our \M.}
  \label{method:model}
\end{figure}
\begin{figure*}[ht]
    \centering
        \begin{subfigure}{0.37\linewidth}
        \includegraphics[width=\linewidth]{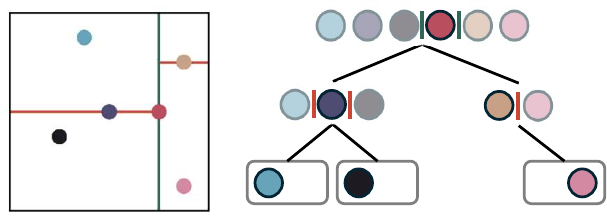}
        \subcaption{KDTree}
        \label{method:orikdt}
      \end{subfigure}%
      \hspace{3mm}
      \begin{subfigure}{0.37\linewidth}
        \includegraphics[width=\linewidth]{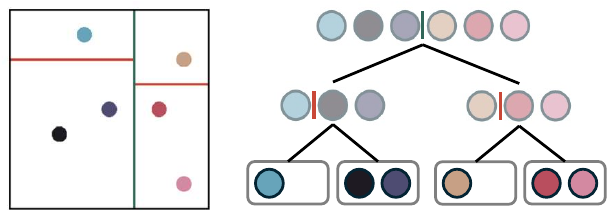}
        \caption{Leaf KDTree}
        \label{method:rikdt}
      \end{subfigure}
      \hspace{3mm}
      \begin{subfigure}{0.18\linewidth}
        \includegraphics[width=\linewidth]{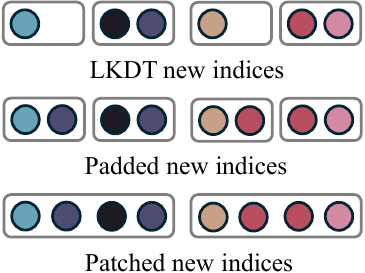}
        \caption{New indices}
        \label{method:idx}
      \end{subfigure}
      \caption{(a)-(b): left part draws the spatial partition using the original and leaf KDTree, and the right part is the corresponding tree. (c): KDT and LKDT are abbreviations of KDTree and leaf KDTree.}
      \label{method:kdts}
\end{figure*}
\section{METHODOLOGY}
In this section, we present our \M~ framework, an effective and efficient solution designed for large-scale traffic forecasting. Figure~\ref{method:model} illustrates the overview of \M, which comprises four primary components: a spatio-temporal embedding module to preprocess traffic data into high-dimensional embeddings, an irregular spatial patching to split the same number of traffic points into patches, a dual attention encoder for extracting spatial information, and a projection decoder for predicting future values. We offer a detailed description of each component in the following.

\subsection{Spatio-Temporal Embedding}
Following previous works~\cite{shao2022spatial,lai2023lightcts}, we adopt a fully-connected layer for each input traffic time series to transform their numerical traffic flow into high-dimensional embeddings. The detailed process of input traffic data $X\in\mathbb{R}^{H\times N}$ can be formulated as follows:
\begin{equation}
    E=W_{(I)}X+b_{(I)}
\end{equation}
where $W_{(I)}\in\mathbb{R}^{d_e\times H}$ and $b_{(I)}\in\mathbb{R}^{d_e}$ are learnable parameters of fully-connected layer. $E\in\mathbb{R}^{N\times d_e}$ is the projected embedding that contains the temporal evolution of traffic. Moreover, we take temporal daily patterns, temporal weekly patterns, and spatial heterogeneous patterns into consideration to improve the distinguishability of points followed by previous methods~\cite{shao2022spatial,liu2023spatio}. For temporal aspect, day-of-week and timeslice-of-day patterns can be stored in the data-driven dictionary $W\in\mathbb{R}^{N_w\times d_w},$ $D\in\mathbb{R}^{N_d\times d_d}$, where $N_w$ and $N_d$ indicate the number of days in a week and the number of timeslices in a day. Therefore, we can use the last timeslice of all points as the index to extract corresponding day-of-week embedding $E_{(w)}\in\mathbb{R}^{N\times d_w}$ and timeslice-of-day embedding $E_{(d)}\in\mathbb{R}^{N\times d_d}$ from dictionaries. Similar to the temporal aspect, we utilize the learnable embedding $E_{(s)}\in\mathbb{R}^{N\times d_s}$ as identities to distinguish points in the dataset. Finally, we concatenate the above embeddings to derive the spatio-temporal embedding:
\begin{equation}
    \tilde{X} = E\vert\vert E_{(w)}\vert\vert E_{(d)}\vert\vert E_{(s)}
\end{equation}
where $\tilde{X}\in\mathbb{R}^{N\times d}$ and $d=d_e+d_w+d_d+d_s$.
\subsection{Irregular Spatial Patching}
\textbf{Motivation.} Spatial propagation is indispensable in improving traffic forecasting performance evaluated by some methods~\cite{liu2023spatio,wen2023diffstg} in addition to spatial distinguishability. This is because vehicles moving on the road will bring real-time traffic changes in the source and destination areas. However, the quadratic complexity of remarkable dynamic spatial modeling methods is unacceptable under current computation resources. Fortunately, we find that spatial information in vision Transformers ~\cite{liu2021swin,wang2023octformer} can be efficiently propagated on the patched input by reducing the number of points involved in attention. The difference between traffic and vision data is that pixels are regularly located in images but traffic points are irregularly distributed on roads, \emph{i.e.}, the same number of pixels can be segmented into patches of the same size, but traffic points can not. Therefore, the main goal of our \M~ is to design a balanced and non-overlapped patching algorithm to reduce computation requirements of performing attention on irregular spatial data.

\noindent\textbf{Leaf KDTree.} As irregular spatial data management is essential for database, geoscience, etc., numerous spatial partitioning algorithms such as KDTree~\cite{sproull1991refinements} and RTree~\cite{guttman1984r} have been proposed. Considering the balance, non-overlapping, and efficiency requirements of partition, we take the simple yet effective KDTree into account to find a solution for evenly dividing irregular traffic data. As shown in Figure~\ref{method:orikdt}, KDTree is a binary spatial tree that uses each internal node as a partitioning hyperplane to split points contained in the node between its two children excluding the hyperplane and is built by recursing on each child node after partitioning, until leaf nodes are reached. Besides, the splitting hyperplane is determined by alternately chosen coordinate axes and the median point of the selected axis. In this paper, locations of latitude and longitude of traffic data are considered as axes to construct the tree. Unfortunately, as illustrated in Figure~\ref{method:orikdt}, we find that hyperplane points in internal nodes of the conventional KDTree are not divided into leaf nodes and may result in irrelevant points being adjacent in the searching order. Therefore, we design a novel leaf KDTree as drawn in Figure~\ref{method:rikdt} to enforce all points stored in leaf nodes, which utilizes the median value as the partitioning hyperplane for internal nodes with an even number of points, and the value between the median point and its left point as the partitioning hyperplane for internal nodes with an odd number of points. Moreover, from the illustration of our leaf KDTree in Figure~\ref{method:rikdt}, we can observe that leaf nodes belonging to the same subtree maintain stronger spatial correlations based on their real-world closer distance, which provides the explainable backtracking for subsequent patching. After constructing leaf KDTree, we conduct the breath first searching on the tree to derive new indices of traffic points according to their searching order, which ensures that leaf nodes belonging to the same subtree are adjacent in the latest index. The entire process based on the latitude $Lat\in\mathbb{R}^{N}$, longitude $Lng\in\mathbb{R}^{N}$, and the capacity $C$ of leaf nodes (leaf nodes in KDTree contain at most $C$ points for a predetermined constant and $C=2$ in Figure~\ref{method:kdts}) can be formulated as follows:
\begin{equation}
\tilde{idx} = BFS(LKDT(Lat,Lng,C))
\end{equation}
where $\tilde{idx}\in\mathbb{R}^{N}$. $LKDT(\cdot)$ and $BFS(\cdot)$ denote the leaf KDTree construction and breath first search operation. 

\noindent \textbf{Padding.} Despite leaf KDTree can provide an equilibrium partition, the number of traffic points $N$ is not necessarily divisible by the capacity $C$, which leads to unfull leaf nodes as shown in Figure~\ref{method:idx}. The inconsistent number destroys the application of patch-based efficient methods. Padding zeros or irrelevant points can mitigate the non-divisible issue yet decrease prediction performance. To make leaf nodes have equaled occupancy and non-self-repeating, we pad the points that are most similar to the unfull leaf nodes from other leaf nodes to reach the maximum capacity, which can confirm the non-overlap patches by the similar time series:
\begin{equation}
    idx,\bar{idx}=Query(LKDT(Lat,Lng,C),CosSim(X,X^T))
\end{equation}
where $Query(\cdot)$ denotes querying most similar points of each unfull leaf node through the Cosine similarity $CosSim(\cdot)$. $idx$ and $\bar{idx}$ indicate locations should be padded in the new index and the corresponding original indices of queried points. Therefore, the padding process can be formulated as follows:
\begin{equation}
     \bar{X} = Pad(idx,\bar{idx},\tilde{X}_{\tilde{idx}})
\end{equation}
where $Pad(\cdot)$ denotes padding queried neighbors into corresponding locations of new indices. Therefore, padded embedding $\bar{X}\in\mathbb{R}^{M\times d}$ have $M$ points and $M=C\times 2^{\lfloor log(N)\rfloor-log(C)}\geq N$.

\noindent \textbf{Patching.} Despite padded leaf nodes have the same number of points, directly using leaf nodes with a large capacity as patches will result in an unbalanced padding issue, \emph{i.e.}, unfull leaf nodes may be padded by points similar to the same point. Recognizing that leaf nodes under the same subtree are strongly correlated and adjacent in new indices, we can first partition points into leaf nodes with a small capacity and then backtrack the tree from leaf nodes to root nodes of subtrees until meet the requirements that the number of the subtree is equal to the number of the pre-defined patch, which mitigates the unbalanced padding issue because unfull leaf nodes are mostly padded by points similar with different points. Concretely, let $P=C\times N_p$ indicate the number of points in a patch and $R$ denotes the number of patches, where $N_p$ is a hyper-parameter that determines how many leaf nodes in a subtree and $M=R\times P$. Notably, $N_p$ can only be a power of $2$ because only leaf nodes that belong to the same subtree have strong spatial locality and our leaf KDTree is a binary tree.

After our balanced and non-overlapped spatial patching, the padded embedding $\bar{X}$ is transformed into a new representation $\mathcal{X}\in\mathbb{R}^{R\times P\times d}$ as the input of the following neural networks.

\subsection{Dual Attention Encoder}
In this section, we present the dual attention encoder to dynamically capture spatial dependencies. For the patched input $\mathcal{X}\in\mathbb{R}^{R\times P\times d}$, $R$ points in the first dimension can be seen as the root nodes of subtrees and $P$ points in the second dimension can be seen points in the subtree. Therefore, \M~ first uses the depth attention on each patch to dynamically extract local spatial information because points in a subtree have stronger correlations. Moreover, as global dependencies are equally essential to the local information in traffic prediction~\cite{fang2022learning}, breadth attention is then adopted on the patch level to learn lossless global knowledge because each point in a root node of the subtree can receive information from points with the same index in other root nodes and these points are mixed with local information after previous depth attention. The dual attention can be interchangeably stacked $L$ layers in the encoder, thus we describe the process of $l$-th layer in the following for simplicity.

\begin{figure}[t]
  \centering
    \includegraphics[width=1.0\linewidth]{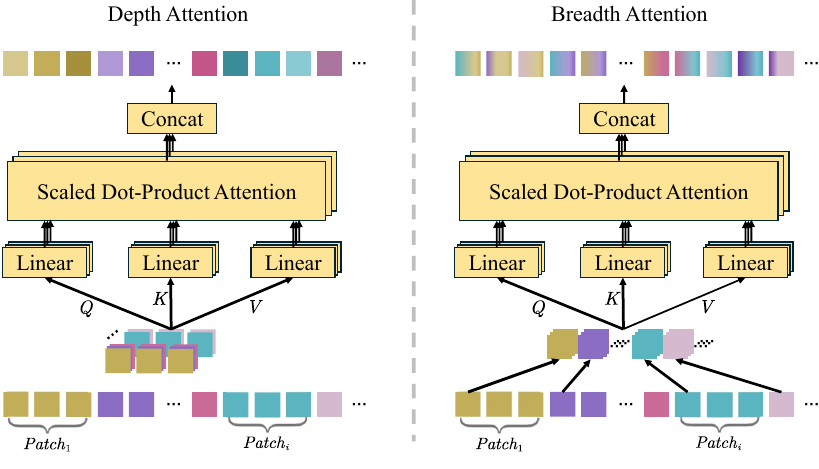}
  \caption{Structure of depth and breadth attention.}
  \label{method:datt}
\end{figure}

\noindent \textbf{Depth Attention.} Depth attention is the multi-head scaled dot-product attention, which is used to capture local spatial information in patches. As shown in Figure~\ref{method:datt}, the query, key, and value are first derived by using learnable linear transformations on the input, which can be formulated as follows:
\begin{equation}
\begin{split}
    Q_{(P)_i}^{(l)} &= W_{(Q)_i}^{(l)}\mathcal{X}^{(l-1)}\\
    K_{(P)_i}^{(l)} &= W_{(K)_i}^{(l)}\mathcal{X}^{(l-1)}\\
    V_{(P)_i}^{(l)} &= W_{(V)_i}^{(l)}\mathcal{X}^{(l-1)}\\
\end{split}
\end{equation}
where $1\le i\le o$ in our multi-head setting. $Q_{(P)_i}^{(l)}, K_{(P)_i}^{(l)}, V_{(P)_i}^{(l)}\in\mathbb{R}^{R\times P\times \frac{d}{o}}$ indicate the query, key, and value of each head in $l$-th layer, and $W_{(Q)_i}^{(l)},W_{(K)_i}^{(l)},W_{(V)_i}^{(l)}\in\mathbb{R}^{d\times \frac{d}{o}}$ are learnable parameters. Notably $\mathcal{X}^{(0)}=\mathcal{X}$ for the encoder. Then we utilize the query and key to dynamically compute point correlations (shape is $\mathbb{R}^{R\times P\times P}$) of each patch, and leverage it to attend spatial information:
\begin{equation}
    A_{(P)_i}^{(l)} = Softmax(\frac{Q_{(P)_i}^{(l)}K_{(P)_i}^{(l)^T}}{\sqrt{d/o}})V_{(P)_i}^{(l)}
\end{equation}
where $Softmax(\cdot)$ normalizes correlations. Finally, multi-head results of the $l$-th layer are concatenated:
\begin{equation}
    \mathcal{\tilde{X}}^{(l)}=(A_{(P)_1}^{(l)}\vert\vert ...\vert\vert A_{(P)_o}^{(l)})W_{(O)}^{(l)}
\end{equation}
where $W_{(O)}^{(l)}\in\mathbb{R}^{d\times d}$ denotes the learnable parameter.

\noindent \textbf{Breadth Attention.} Breadth attention is also the multi-head scaled dot-product attention, which is used to learn global spatial knowledge at the patch level. Similarity, the query, key, and value $Q_{(R)_i}^{(l)}$, $K_{(R)_i}^{(l)}$, $V_{(R)_i}^{(l)}\in\mathbb{R}^{P\times R\times \frac{d}{o}}$ are first derived by using learnable linear transformations on the output of depth attention. Then query and key are used to dynamically compute patch correlations (shape is $\mathbb{R}^{P\times R\times R}$) of each index in patches. Finally, the output of the $l$-th layer $\mathcal{X}^{(l)}\in\mathbb{R}^{R\times P\times d}$ is derived by using the calculated correlations to attend global information.

\subsection{Projection Decoder}
In this section, we aim to predict the future traffic through the spatial information interacted output $\mathcal{X}^{(L)}\in\mathbb{R}^{R\times P\times d}$ of the dual attention encoder. We first unpatch the output to $\bar{X}^{(L)}\in\mathbb{R}^{M\times d}$ through performing depth first search operation on each root node, and unpad leaf nodes to consist with the input shape and index:
\begin{equation}
    \tilde{Y}_{\tilde{idx}} = UnPad(idx,\bar{X}^{(L)})
\end{equation}
where $UnPad(\cdot)$ denotes removing points on the unpadded locations and $\tilde{Y}\in\mathbb{R}^{N\times d}$ is the representation with original index. Finally, a fully connected layer is adopted to project the historical representation into the future:
\begin{equation}
    \hat{Y}=W_{(D)}\tilde{Y}+b_{(D)}
\end{equation}
where $W_{(D)}\in\mathbb{R}^{F\times d},$ $b_{(D)}\in\mathbb{R}^{F}$ are learnable parameters and $\hat{Y}\in\mathbb{R}^{F\times N}$ indicates predicted traffic. During the training stage, we use $\hat{Y}$ and $Y$ to compute L1 loss as the objective function to guide the learning of \M.

\subsection{Complexity Analysis}
The leaf KDTree takes $O(Nlog(N))$ complexity to construct the balanced binary tree based on the median hyperplane~\cite{de2008orthogonal}, which can be quickly done in the pre-processing stage. In the dual attention encoder, the cost of depth and breadth attention is respectively $O(RP^2d)$ and $O(PR^2d)$. Therefore, the dominant complexity of \M~ is $O(max(P,R)Md)$, which requires less time than quadratic dynamic spatial modeling methods because $P\ll N$, $R\ll N$, and $M\approx N$ in \M.

\begin{table}[ht]
    \centering
    \vspace{-5pt}
    \caption{Dataset statistics.}
    \vspace{-5pt}
    \resizebox{1.0\linewidth}{!}{\begin{tabular}{lcccc}
    \toprule
    Datasets & \#Points & \#Samples & \#TimeSlices & Timespan\\
    \midrule
    SD & 716 & 25M & 35040 & 01/01/2019-12/31/2019\\
    GBA & 2352 & 82M  & 35040  & 01/01/2019-12/31/2019\\
    GLA & 3834 & 134M & 35040 & 01/01/2019-12/31/2019\\
    CA & 8600 & 301M & 35040 & 01/01/2019-12/31/2019\\
    \bottomrule
    \end{tabular}}
    \label{exp:data}
    \vspace{-10pt}
\end{table}

\section{Experiments}
The goal of this section is to address the following five pivotal research questions by conducting comprehensive experiments on four large-scale traffic datasets.
\begin{itemize}[leftmargin=*]
    \item \textbf{RQ1:} How does \M~ perform when compared to current approaches in large-scale traffic forecasting?
    \item \textbf{RQ2:} What contributions do the main components of \M?
    \item \textbf{RQ3:} How efficient is \M~ in large-scale datasets?
    \item \textbf{RQ4:} Does \M~ output reasonable correlations?
    \item \textbf{RQ5:} How do the essential hyper-parameters impact \M?
\end{itemize}

\begin{table*}[t]
\aboverulesep=0ex
\belowrulesep=0ex
\centering
\caption{Large-scale traffic forecasting performance comparison of our \M~and baselines. \redfont{Redfont} indicates the best performance and \bluefont{bluefont} denotes the second best performance.}
\resizebox{1.0\linewidth}{!}{
\begin{tabular}{c|c|ccc|ccc|ccc|ccc}
\toprule[1.5pt]
\multicolumn{1}{c|}{\multirow{2}{*}{Datasets}} & \multicolumn{1}{c|}{\multirow{2}{*}{Methods}} & \multicolumn{3}{c|}{Horizon 3} & \multicolumn{3}{c|}{Horizon 6} & \multicolumn{3}{c|}{Horizon 12} & \multicolumn{3}{c}{Average}\\
\cmidrule{3-14}
    &  & MAE & RMSE & MAPE ($\%$) & MAE & RMSE & MAPE ($\%$) & MAE & RMSE & MAPE ($\%$) & MAE & RMSE & MAPE ($\%$)\\
\midrule
\midrule
\multirow{11}{*}{SD}
& STID & 15.15 & 25.29 & 9.82 & 17.95 & 30.39 & 11.93 & 21.82 & 38.63 & 15.09 & 17.86 & 31.00 & 11.94\\
\cdashline{2-14}
& GWNET & 15.24 & 25.13 & 9.86 & 17.74 & 29.51 & 11.70 & \bluefont{21.56} & \bluefont{36.82} & 15.13 & \bluefont{17.74} & 29.62 & 11.88\\
& AGCRN &  15.71 & 27.85 & 11.48 & 18.06 & 31.51 & 13.06 & 21.86 & 39.44 & 16.52 & 18.09 & 32.01 & 13.28\\
& STGODE & 16.75 & 28.04 & 11.00 & 19.71 & 33.56 & 13.16 & 23.67 & 42.12 & 16.58 & 19.55 & 33.57 & 13.22\\
& RPMixer & 18.54 & 30.33 & 11.81 & 24.55 & 40.04 & 16.51 & 35.90 & 58.31 & 27.67 & 25.25 & 42.56 & 17.64\\
\cdashline{2-14}
& DSTAGNN & 18.13 & 28.96 & 11.38 & 21.71 & 34.44 & 13.93 & 27.51 & 43.95 & 19.34 & 21.82 & 34.68 & 14.40\\
& D2STGNN & \bluefont{14.92} & \bluefont{24.95} & \bluefont{9.56} & \bluefont{17.52} & \bluefont{29.24} & \bluefont{11.36} & 22.62 & 37.14 & \bluefont{14.86} & 17.85 & \bluefont{29.51} & \bluefont{11.54}\\
& DGCRN & 15.34 & 25.35 & 10.01 & 18.05 & 30.06 & 11.90 & 22.06 & 37.51 & 15.27 & 18.02 & 30.09 & 12.07\\
& STWave & 15.80 & 25.89 & 10.34 & 18.18 & 30.03 & 11.96 & 21.98 & 36.99 & 15.30 & 18.22 & 30.12 & 12.20\\
& BigST & 16.42 & 26.99 & 10.86 & 18.88 & 31.60 & 13.24 & 23.00 & 38.59 & 15.92 & 18.80 & 31.73 & 12.91\\
\cmidrule{2-14}
& \M & \redfont{14.53} & \redfont{24.34} & \redfont{9.22} & \redfont{16.86} & \redfont{28.63} & \redfont{11.11} & \redfont{20.66} & \redfont{36.27} & \redfont{14.72} & \redfont{16.90} & \redfont{29.27} & \redfont{11.23}\\
\midrule
\midrule
\multirow{11}{*}{GBA}
& STID & \bluefont{17.36} & 29.39 & 13.28 & \bluefont{20.45} & 34.51 & 16.03 & \bluefont{24.38} & 41.33 & 19.90 & \bluefont{20.22} & 34.61 & 15.91\\
\cdashline{2-14}
& GWNET & 17.85 & 29.12 & 13.92 & 21.11 & \bluefont{33.69} & 17.79 & 25.58 & 40.19 & 23.48 & 20.91 & \bluefont{33.41} & 17.66\\
& AGCRN & 18.31 & 30.24 & 14.27 & 21.27 & 34.72 & 16.89 & 24.85 & \bluefont{40.18} & 20.80 & 21.01 & 34.25 & 16.90\\
& STGODE & 18.84 & 30.51 & 15.43 & 22.04 & 35.61 & 18.42 & 26.22 & 42.90 & 22.83 & 21.79 & 35.37 & 18.26\\
& RPMixer & 20.31 & 33.34 & 15.64 & 26.95 & 44.02 & 22.75 & 39.66 & 66.44 & 37.35 & 27.77 & 47.72 & 23.87\\
\cdashline{2-14}
& DSTAGNN & 19.73 & 31.39 & 15.42 & 24.21 & 37.70 & 20.99 & 30.12 & 46.40 & 28.16 & 23.82 & 37.29 & 20.16\\
& D2STGNN & 17.54 & \bluefont{28.94} & \redfont{12.12} & 20.92 & 33.92 & \bluefont{14.89} & 25.48 & 40.99 & \bluefont{19.83} & 20.71 & 33.65 & \bluefont{15.04}\\
& DGCRN & 18.02 & 29.49 & 14.13 & 21.08 & 34.03 & 16.94 & 25.25 & 40.63 & 21.15 & 20.91 & 33.83 & 16.88\\
& STWave & 17.95 & 29.42 & 13.01 & 20.99 & 34.01 & 15.62 & 24.96 & 40.31 & 20.08 & 20.81 & 33.77 & 15.76\\
& BigST & 18.70 & 30.27 & 15.55 & 22.21 & 35.33 & 18.54 & 26.98 & 42.73 & 23.68 & 21.95 & 35.54 & 18.50\\
\cmidrule{2-14}
& \M & \redfont{16.81} & \redfont{28.71} & \bluefont{12.25} & \redfont{19.68} & \redfont{33.09} & \redfont{14.51} & \redfont{23.49} & \redfont{39.23} & \redfont{18.93} & \redfont{19.50} & \redfont{33.16} & \redfont{14.64}\\
\midrule
\midrule
\multirow{9}{*}{GLA}
& STID & \bluefont{16.54} & 27.73 & \bluefont{10.00} & \bluefont{19.98} & 34.23 & \bluefont{12.38} & \bluefont{24.29} & 42.50 & \bluefont{16.02} & \bluefont{19.76} & 34.56 & \bluefont{12.41}\\
\cdashline{2-14}
& GWNET & 17.28 & \bluefont{27.68} & 10.18 & 21.31 & 33.70 & 13.02 & 26.99 & 42.51 & 17.64 & 21.20 & 33.58 & 13.18\\
& AGCRN & 17.27 & 29.70 & 10.78 & 20.38 & 34.82 & 12.70 & 24.59 & 42.59 & 16.03 & 20.25 & 34.84 & 12.87\\
& STGODE & 18.10 & 30.02 & 11.18 & 21.71 & 36.46 & 13.64 & 26.45 & 45.09 & 17.60 & 21.49 & 36.14 & 13.72\\
& RPMixer & 19.94 & 32.54 & 11.53 & 27.10 & 44.87 & 16.58 & 40.13 & 69.11 & 27.93 & 27.87 & 48.96 & 17.66\\
\cdashline{2-14}
& DSTAGNN & 19.49 & 31.08 & 11.50 & 24.27 & 38.43 & 15.24 & 30.92 & 48.52 & 20.45 & 24.13 & 38.15 & 15.07\\
& STWave & 17.48 & 28.05 & 10.06 & 21.08 & \bluefont{33.58} & 12.56 & 25.82 & \bluefont{41.28} & 16.51 & 20.96 & \bluefont{33.48} & 12.70\\
& BigST & 18.38 & 29.40 & 11.68 & 22.22 & 35.53 & 14.48 & 27.98 & 44.74 & 19.65 & 22.08 & 36.00 & 14.57\\
\cmidrule{2-14}
& \M & \redfont{15.84} & \redfont{26.34} & \redfont{9.27} & \redfont{19.06} & \redfont{31.85} & \redfont{11.30} & \redfont{23.32} & \redfont{39.64} & \redfont{14.60} & \redfont{18.96} & \redfont{32.33} & \redfont{11.44}\\
\midrule
\midrule
\multirow{7}{*}{CA}
& STID & \bluefont{15.51} & \bluefont{26.23} & \bluefont{11.26} & \bluefont{18.53} & 31.56 & \bluefont{13.82} & \bluefont{22.63} & 39.37 & \bluefont{17.59} & \bluefont{18.41} & 32.00 & \bluefont{13.82}\\
\cdashline{2-14}
& GWNET & 17.14 & 27.81 & 12.62 & 21.68 & 34.16 & 17.14 & 28.58 & 44.13 & 24.24 & 21.72 & 34.20 & 17.40\\
& STGODE & 17.57 & 29.91 & 13.91 & 20.98 & 36.62 & 16.88 & 25.46 & 45.99 & 21.00 & 20.77 & 36.60 & 16.80\\
& RPMixer & 18.18 & 30.49 & 12.86 & 24.33 & 41.38 & 18.34 & 35.74 & 62.12 & 30.38 & 25.07 & 44.75 & 19.47\\
\cdashline{2-14}
& STWave & 16.77 & 26.98 & 12.20 & 18.97 & \bluefont{30.69} & 14.40 & 25.36 & \bluefont{38.77} & 19.01 & 19.69 & \bluefont{31.58} & 14.58\\
& BigST & 17.15 & 27.92 & 13.03 & 20.44 & 33.16 & 15.87 & 25.49 & 41.09 & 20.97 & 20.32 & 33.45 & 15.91\\
\cmidrule{2-14}
& \M & \redfont{14.69} & \redfont{24.82} & \redfont{10.51} & \redfont{17.41} & \redfont{29.43} & \redfont{12.83} & \redfont{21.20} & \redfont{36.13} & \redfont{16.00} & \redfont{17.35} & \redfont{29.79} & \redfont{12.79}\\
\bottomrule[1.5pt]
\end{tabular}
}
\label{exp:main}
\end{table*}

\subsection{Experimental Setup}

\subsubsection{Datasets}
We conduct experiments on four large-scale datasets SD, GBA, GLA, and CA as introduced in LargeST~\cite{liu2024largest}. Following previous settings, we not only chronologically split each dataset into train, validation, and test sets with a ratio of 6:2:2 but also utilize continuous $24$ time slices as samples to perform traffic forecasting with the historical $12$ time slices as the input and the future $12$ time slices as the output. Detailed statistics of these datasets are shown in Table~\ref{exp:data}.

\subsubsection{Baselines.}
We compare 10 advanced baselines in this paper with our \M~. (i) The non-spatial modeling-based STID~\cite{shao2022spatial}, which uses identity spatio-temporal embeddings in fully-connected layers to forecast traffic. (ii) In the category of static spatial-based methods, we select GWNET~\cite{wu2019graph}, AGCRN~\cite{bai2020adaptive}, STGODE~\cite{fang2021spatial}, and RPMixer~\cite{yeh2024rpmixer}. They combine various static GNNs with temporal networks for traffic forecasting. (iii) Baselines in the dynamic spatial-based category include, DSTAGNN~\cite{lan2022dstagnn}, D2STGNN~\cite{shao2022decoupled}, DGCRN~\cite{li2023dynamic}, low-rank STWave~\cite{fang2023spatio}, and linear BigST~\cite{han2024bigst}. They dynamically reveal spatial correlations at different time periods for traffic forecasting. 

\subsubsection{Evaluation Metrics.} 
We utilize diverse evaluation criteria from performance and efficiency aspects for a comprehensive comparison. For the performance aspect: we utilize three commonly adopted numerical metrics to assess the performance of predicted traffic time series, \emph{i.e.}, mean absolute error (MAE), root mean squared error (RMSE), and mean absolute percentage error (MAPE). For the efficiency aspect: the measurement of the model's efficiency is based on the wall-clock time, and the memory needs of models are revealed by the batch size in the training phase.

\subsubsection{Implementation Details.}
During training, \M~ is optimized by the AdamW optimizer with a learning rate of $0.002$ and weight decay of $0.0001$, and the training epoch is set to 50 for all datasets. Moreover, the learning rate is halved at $2$, $35$, and $40$ epochs. To better reproduce our model, we summarize all the default hyper-parameters as follows. The dimension of input projection in SD, GBA, GLA, and CA datasets is set to $128$, $128$, $64$, and $64$. The dimension of day-of-week embedding, timeslice-of-day embedding, and spatial embedding is all set to $32$. The number of segmented patches of CA, GLA, GBA, and SD is set to $512$, $64$, $16$, and $16$. The number of points in a leaf node is set as $3$, $2$, $2$, and $2$ for CA, GLA, GBA, and SD. The number of attention layers is set to $5$. Besides, all the experiments are implemented in PyTorch with the NVIDIA RTX A6000 48GB GPU. The source code of \M~ is available at: https://github.com/LMissher/PatchSTG.

\subsection{Performance Comparisons (RQ1)}
\label{Main Results}
Table~\ref{exp:main} showcases the MAE, RMSE, and MAPE of traffic forecasting across all methods on four large-scale datasets except for failure to run with the smallest batch size of $1$. The performance on the horizon $3$, horizon $6$, horizon $12$, and the average of the whole $12$ horizons are reported. To ensure a fair comparison, we follow official configurations of baselines, with the only adjustment of fixes input length to $12$. Therefore baselines in our paper may show slight variations compared to the original results.

\textbf{Advantages of Distinguishability.} Among the baselines concerned, identity embedding-based non-spatial STID, shows a significant lead over spatial modeling methods on larger datasets such as the CA dataset. This phenomenon underscores the advantages of learning heterogeneous representations of different points to avoid over-smoothing in spatial message passing.

\textbf{Advantages of Dynamic Spatial Modeling.} As shown in Table~\ref{exp:main}, D2STGNN exhibits remarkable superiority, especially on the SD dataset when compared to non-spatial and static spatial-based methods. This appearance is owing to the dynamic spatial modeling. Despite point-to-point D2STGNN achieving great performance in small-scale datasets, it is still limited by the quadratic complexity in large-scale GLA and CA datasets.

\textbf{Advantages of Explicit Spatial Aggregation.} In large-scale datasets, in contrast to linear-based efficient dynamic spatial modeling method BigST, low rank-based STWave demonstrates clear advantages on all datasets under spatial reduction caused information loss, attributed to the implicit spatial correlations in linear-based methods are failed to correctly normalize.

\textbf{Consistent Performance Superiority.} Drawing on the aforementioned components and our irregular spatial patching, we introduce \M, an efficient Transformer framework that achieves state-of-the-art performance on all datasets as evidenced in Table~\ref{exp:main}.
\begin{table*}[t]
    \caption{Ablation study of \M~on average results of large-scale traffic datasets. \textbf{Bold}: best performance.}
    \aboverulesep=0ex
    \belowrulesep=0ex
    \centering
    \begin{tabular}{|l|ccc|ccc|ccc|ccc|}
    \toprule[0.6pt]
        Dataset& \multicolumn{3}{c|}{SD} & \multicolumn{3}{c|}{GBA} & \multicolumn{3}{c|}{GLA} & \multicolumn{3}{c|}{CA}\\
    \midrule[0.5pt]
         Metric & MAE & RMSE & MAPE (\%) & MAE & RMSE & MAPE (\%) & MAE & RMSE & MAPE (\%) & MAE & RMSE & MAPE (\%)\\
    \midrule[0.5pt]
        w/o FGGC & 17.37 & 29.44 & 11.45 & 19.72 & 33.28 & 15.51 & 19.49 & 32.95 & 11.69 & 17.63 & 30.04 & 13.05\\
        w/o Depth & 17.18 & 29.91 & 11.28 & 19.63 & 33.33 & 15.11 & 19.38 & 33.05 & 11.73 & 17.68 & 30.32 & 13.09\\
        w/o Breadth & 17.70 & 30.55 & 11.46 & 19.90 & 33.44 & 15.28 & 19.65 & 33.66 & 12.18 & 18.04 & 30.74 & 13.17\\
    \midrule[0.5pt]
        w/ PadDis & 17.02 & 29.42 & 11.34 & 19.60 & 33.32 & 15.03 & 19.17 & 32.80 & 11.63 & 17.50 & 30.14 & 13.00\\
        w/o PadSim & 17.00 & 29.43 & 11.25 & 19.57 & 33.34 & 15.32 & 19.57 & 33.18 & 11.80 & 17.87 & 30.52 & 13.11\\
    \midrule[0.5pt]
        w/ METIS & 18.00 & 30.69 & 11.72 & 19.91 & 33.38 & 15.16 & 19.63 & 33.57 & 11.86 & 18.02 & 30.53 & 13.02\\
        w/ KMeans & 18.07 & 30.76 & 11.87 & 20.16 & 33.93 & 15.87 & 19.85 & 34.07 & 11.90 & 18.26 & 31.08 & 13.05\\
        w/o LKDT & 17.58 & 29.38 & 11.45 & 20.14 & 34.18 & 15.91 & 19.88 & 34.47 & 12.31 & 18.27 & 30.97 & 13.00\\
    \midrule[0.5pt]
        \M & \textbf{16.90} & \textbf{29.27} & \textbf{11.23} & \textbf{19.50} & \textbf{33.16} & \textbf{14.64} & \textbf{18.96} & \textbf{32.33} & \textbf{11.44} & \textbf{17.35} & \textbf{29.79} & \textbf{12.79}\\
    \bottomrule[0.6pt]
    \end{tabular}
    \label{exp:abl}
\end{table*}

\begin{table*}[t]
    \caption{Efficiency comparisons on large-scale traffic datasets. BS: batch size. Train: training time (in seconds) per epoch. Infer: inference time (in seconds). Total: total training time (in hours). Note that $Total=Train\times Epoches$ and - indicates out of memory.}
    \aboverulesep=0ex
    \belowrulesep=0ex
    \centering
    \resizebox{1.0\linewidth}{!}{\begin{tabular}{|c|cccc|cccc|cccc|cccc|}
    \toprule[0.6pt]
        \multicolumn{1}{|c|}{\multirow{2}{*}{Methods}} & \multicolumn{4}{c|}{SD} & \multicolumn{4}{c|}{GBA} & \multicolumn{4}{c|}{GLA} & \multicolumn{4}{c|}{CA}\\
    \cmidrule{2-17}
        & BS & Train & Infer & Total & BS & Train & Infer & Total & BS & Train & Infer & Total & BS & Train & Infer & Total\\
    \midrule[0.5pt]
        DSTAGNN & 64 & 240 & 23 & 7 & 27 & 1959 & 171 & 53 & 10 & 5241 & 467 & 120 & - & - & - & -\\
        D2STGNN & 45 & 563 & 69 & 14 & 4 & 5885 & 796 & 148 & - & - & - & - & - & - & - & -\\
        DGCRN & 64 & 430 & 76 & 14 & 12 & 4461 & 605 & 138 & - & - & - & - & - & - & - & -\\
        STWave & 64 & 259 & 33 & 7 & 36 & 926 & 120 & 26 & 22 & 1483 & 179 & 41 & 8 & 3641 & 437 & 101\\
    \midrule[0.5pt]
        \M & 64 & 64 & 6 & 1 & 64 & 262 & 30 & 4 & 64 & 295 & 27 & 4 & 32 & 981 & 93 & 14\\
    \cdashline{1-17}
        Improvements & 0$\times$ & 3.8$\times$ & 3.8$\times$ & 7$\times$ & 1.8$\times$ & 1.8$\times$ & 4$\times$ & 6.5$\times$ & 2.9$\times$ & 5$\times$ & 6.6$\times$ & 10$\times$ & 4$\times$ & 3.7$\times$ & 4.7$\times$ & 7.2$\times$\\
    \bottomrule[0.6pt]
    \end{tabular}}
    \label{exp:eff}
\end{table*}
\subsection{Ablation Study (RQ2)}
\label{Ablation Study}
In this section, ablation experiments are conducted on four datasets using four variants of \M~ to address RQ2. These four variants are listed below:
\begin{itemize}[leftmargin=*]
    \item "w/o FGGC": \M~ fuse all points in the same patch into a single patch point during the breadth attention.
    \item "w/o Depth": \M~ removes the depth attention and only global spatial correlations are modeled.
    \item "w/o Breadth": \M~ removes the breadth attention and only local spatial correlations are modeled.
    \item "w/ PadDis": \M~ pads points with closest distance into unfull patches.
    \item "w/o PadSim": This variant only pads zero constants but not other points into unfull leaf nodes.
    \item "w/ METIS": This variant uses the balanced graph partition algorithm METIS to replace our Leaf KDTree.
    \item "w/ KMeans": This variant utilizes the unbalanced clustering algorithm KMeans to substitute our Leaf KDTree.
    \item "w/o LKDT": \M~ no longer equips the leaf KDTree, \emph{i.e.}, dual attention is conducted on the original input.
\end{itemize}
According to the results illustrated in Table~\ref{exp:abl}, the following observations can be found.

\textbf{Benefits Brought by KDTree.} Experimental results of "w/o LKDT" on all datasets show a huge drop in prediction performance, recommending that the most important in \M~ is not dynamically modeling spatial dependencies on irrelevant points in the patch but passing spatial messages on adjacent points in the patch. Moreover, the performance of STID and "w/o LKDT" on the GLA dataset further verifies that our leaf KDTree is the key component to model local spatial information. Besides, we test some methods that can also segment the traffic data into multiple patches. The conventional graph cluster methods such as KMeans not only achieve poor performance as "w/ KMeans" but also fail to derive balanced patching (the maximum and minimum patch size of KMeans on the SD dataset is 96 and 6). Though the graph partition method METIS can approximately segment traffic data into balanced patches, it cannot derive balance padded non-overlap patches because the recursive merge operation in our leaf KDTree is not supported, and thus performs badly as "w/ METIS".

\textbf{Effectiveness of Non-Overlap Padding.} The experiments of "w/ PadDis" and "w/o PadSim" reflect that, in all datasets, replacing similar points with zeros or neighbored points to pad unfull leaf nodes results in a drop in performance. This finding underscores the effectiveness of our non-overlap padding from the temporal view and backtracking merging for balanced padding.

\textbf{Effectiveness of Dual Attention.} As observed by the decreased performance in "w/o Depth," depth attention is essential in \M~ to capture local spatial dependencies. Similar to the local information, global spatial interactions are also indispensable in spatial modeling as the performance drop of "w/o Breadth". Besides, compared with the "w/o FGGC" performance, our non-fused fine-grained global modeling achieves better results because it preserves diverse global information and thus our \M~ is fidelity.

\subsection{Efficiency Comparisons (RQ3)}
To evaluate the efficiency of our \M, we present the training speed, inference speed, and batch size comparisons among our \M~ and previous dynamic spatial modeling methods with explicit spatial correlations, including DSTAGNN, D2STGNN, DGCRN, and STWave. As depicted in Table~\ref{exp:eff}, \M~ attains the fastest speed and boasts the most efficient GPU memory utilization across all datasets, especially achieving up to $10\times$ and $4\times$ improvements in speed and memory on large-scale GLA and CA datasets. While low-rank-based STWave also excels other methods in speed on large-scale datasets, the worse performance of STWave on small datasets compared to quadratic complexity-based DSTAGNN suggests that low-rank-based methods are less general than our \M~ under different settings.

\subsection{Visualization (RQ4)}
\begin{figure}[t]
  \centering
\includegraphics[width=0.82\linewidth]{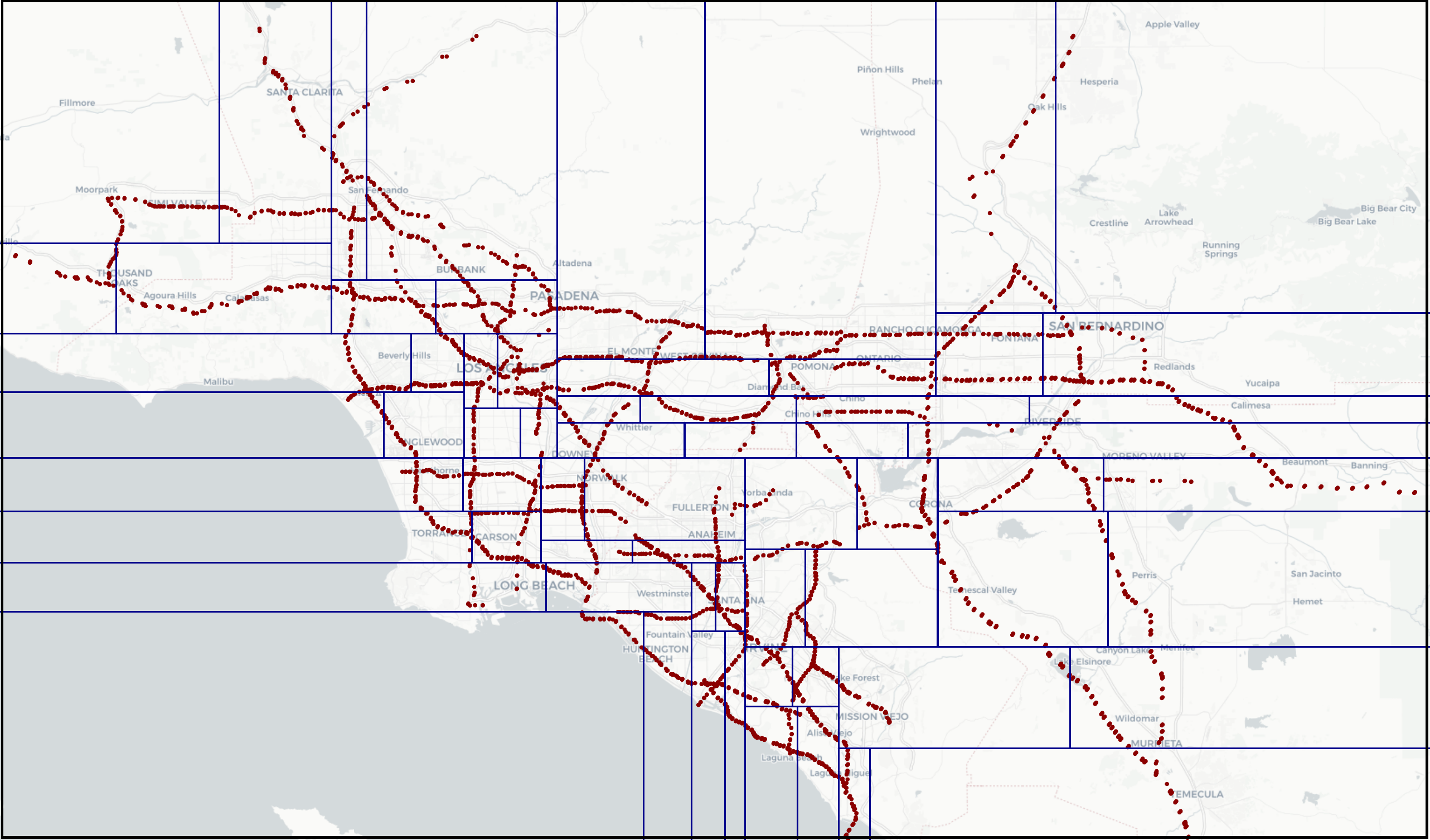}
  \caption{Leaf KDTree on the GLA dataset.}
  \label{exp:vis}
\end{figure}

\textbf{Interpretability.}
To verify the interpretability of our \M, we visualize the evenly segmented GLA dataset using our leaf KDTree in Figure~\ref{exp:vis}. We can observe that the real-world adjacent points are obviously divided into the same leaf node to maintain spatial locality, which can give explainable spatial partition compared with low-rank-based dynamic spatial modeling methods. Moreover, our \M~ can explicitly show the learned local and global spatial correlations corresponding to their real-world locations on the segmented map, which is interpretable compared with linear-based dynamic spatial modeling methods.

\begin{figure}[t]
    \centering
    \begin{subfigure}{0.35\linewidth}
        \includegraphics[width=\linewidth]{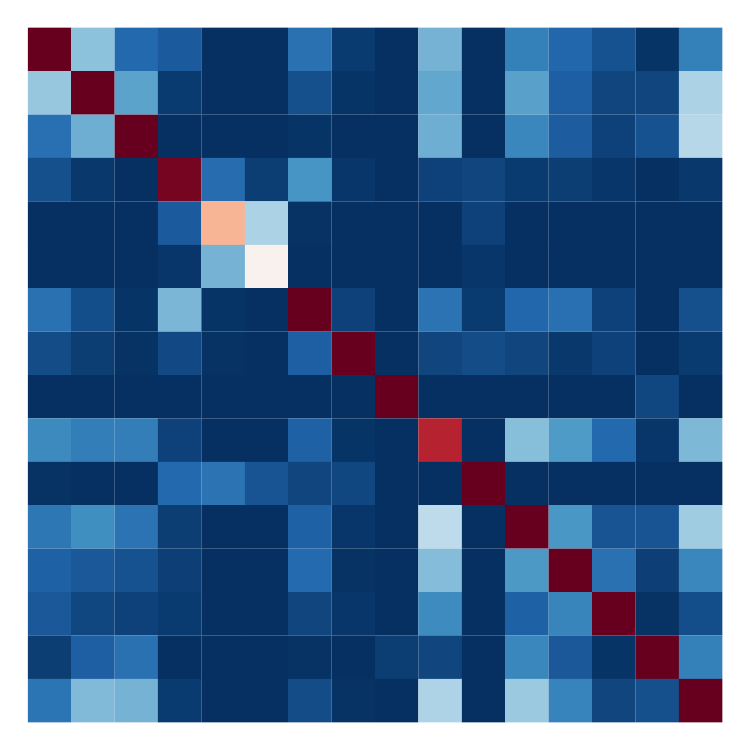}
        \caption{Correlations on 23$^{th}$}
        \label{exp:23}
      \end{subfigure}
      \hspace{10mm}
      \begin{subfigure}{0.35\linewidth}
        \includegraphics[width=\linewidth]{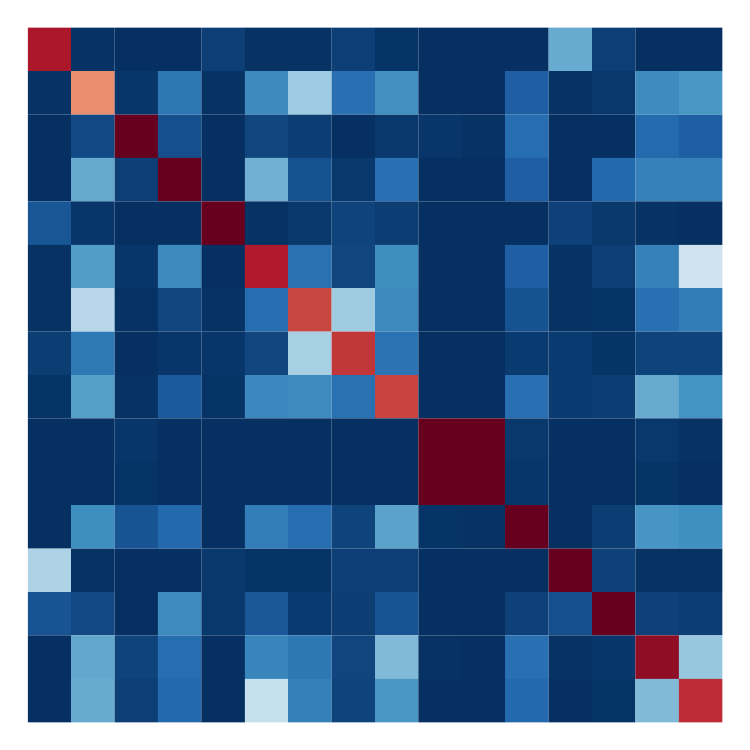}
        \caption{Correlations on 56$^{th}$}
        \label{exp:42}
      \end{subfigure}%
      \caption{Learned patch-level correlations.}
      \label{exp:corr}
\end{figure}

\noindent\textbf{Fidelity.}
To show our framework of learning global spatial knowledge without information loss, we conduct a case study on the GLA dataset, \emph{i.e.}, we visualize the learned patch correlations of points with indices 23 and 56 in the patch. As shown in Figure~\ref{exp:corr}, the global correlations learned by the breadth attention are diverse for different indices in the patch, which follows the heterogeneity of traffic points. Therefore, our \M~ is fidelity compared with low-rank-based methods because they can only reflect same patch correlations for all points in the patch.

\subsection{Hyper-parameter Study (RQ5)}
\label{Hyper-parameter Study}
\begin{figure}[t]
    \centering
    \begin{subfigure}{0.45\linewidth}
        \includegraphics[width=\linewidth]{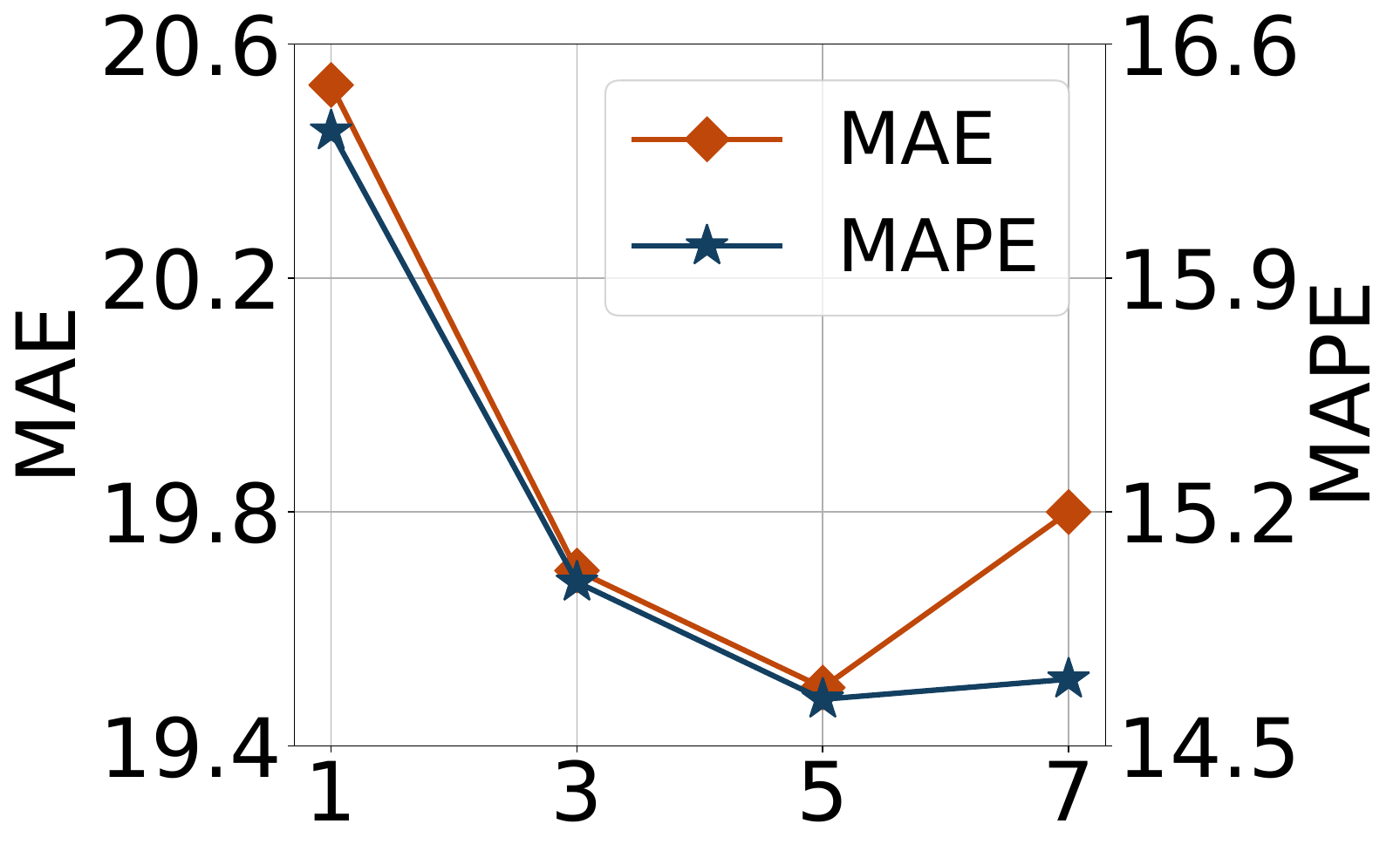}
        \caption{Layers on GBA}
        \label{exp:gbalayers}
      \end{subfigure}
      \hspace{2mm}
      \begin{subfigure}{0.45\linewidth}
        \includegraphics[width=\linewidth]{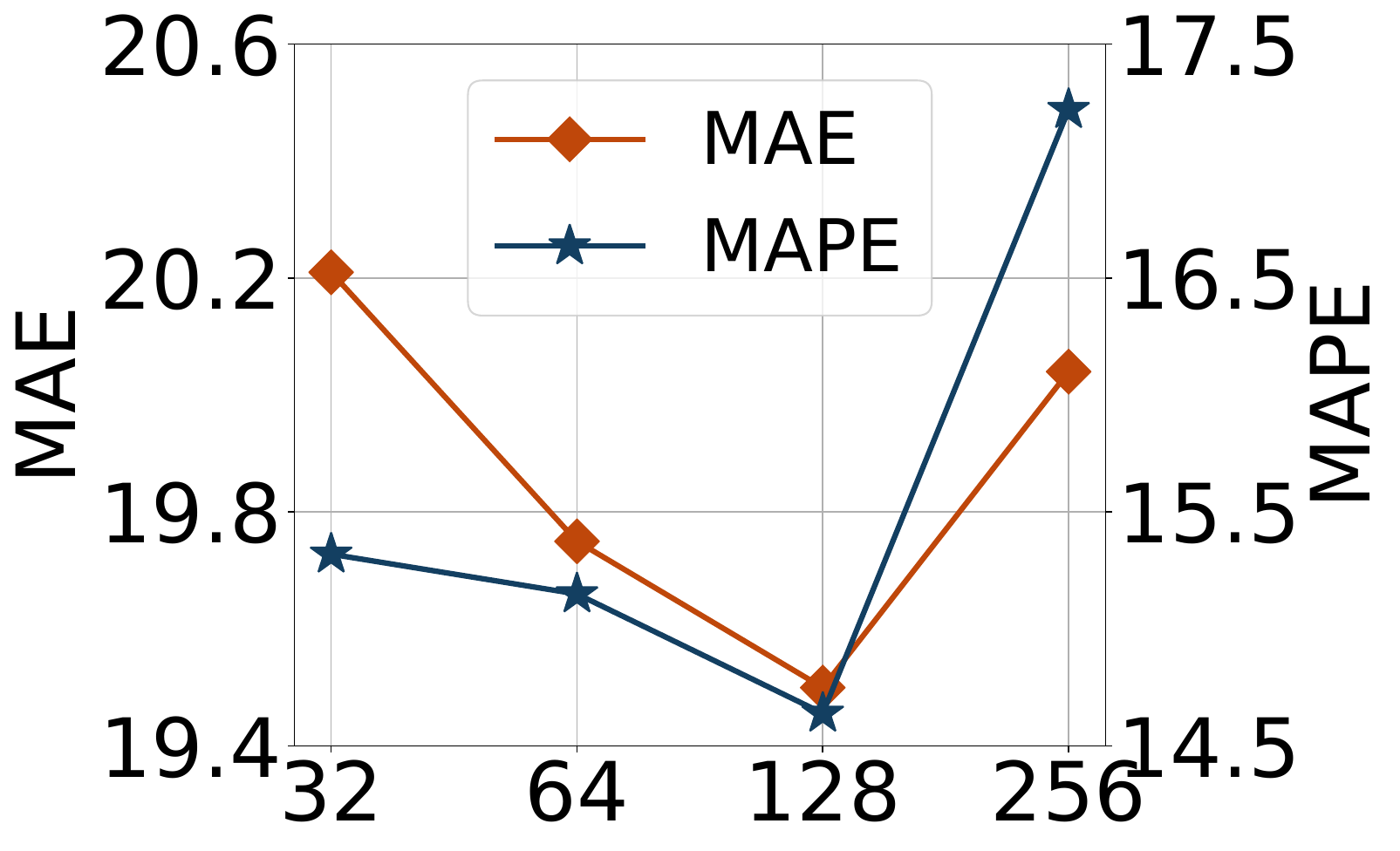}
        \caption{Dimensions on GBA}
        \label{exp:gbadims}
      \end{subfigure}%

      \begin{subfigure}{0.45\linewidth}
        \includegraphics[width=\linewidth]{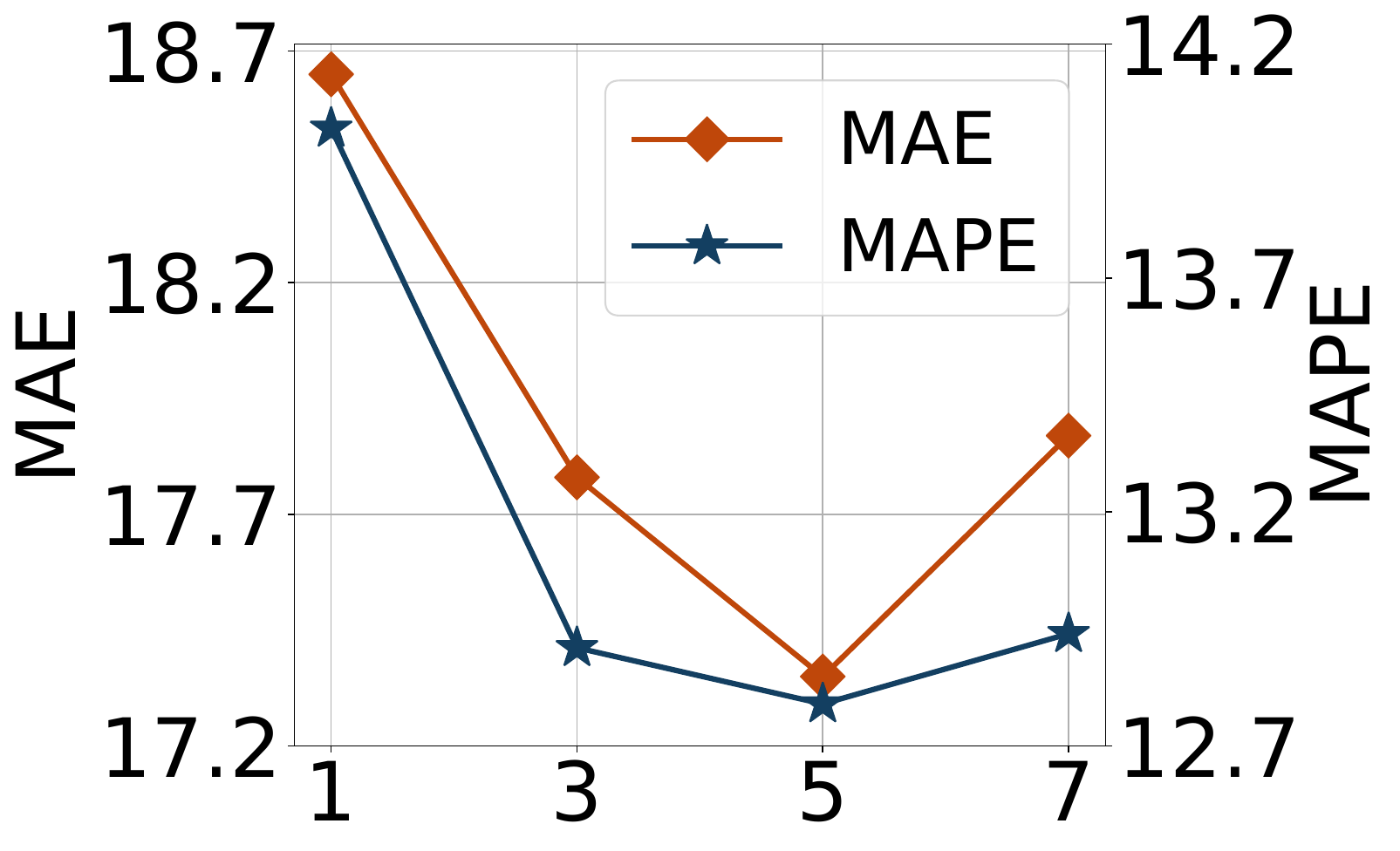}
        \caption{Layers on CA}
        \label{exp:calayers}
      \end{subfigure}
      \hspace{2mm}
      \begin{subfigure}{0.45\linewidth}
        \includegraphics[width=\linewidth]{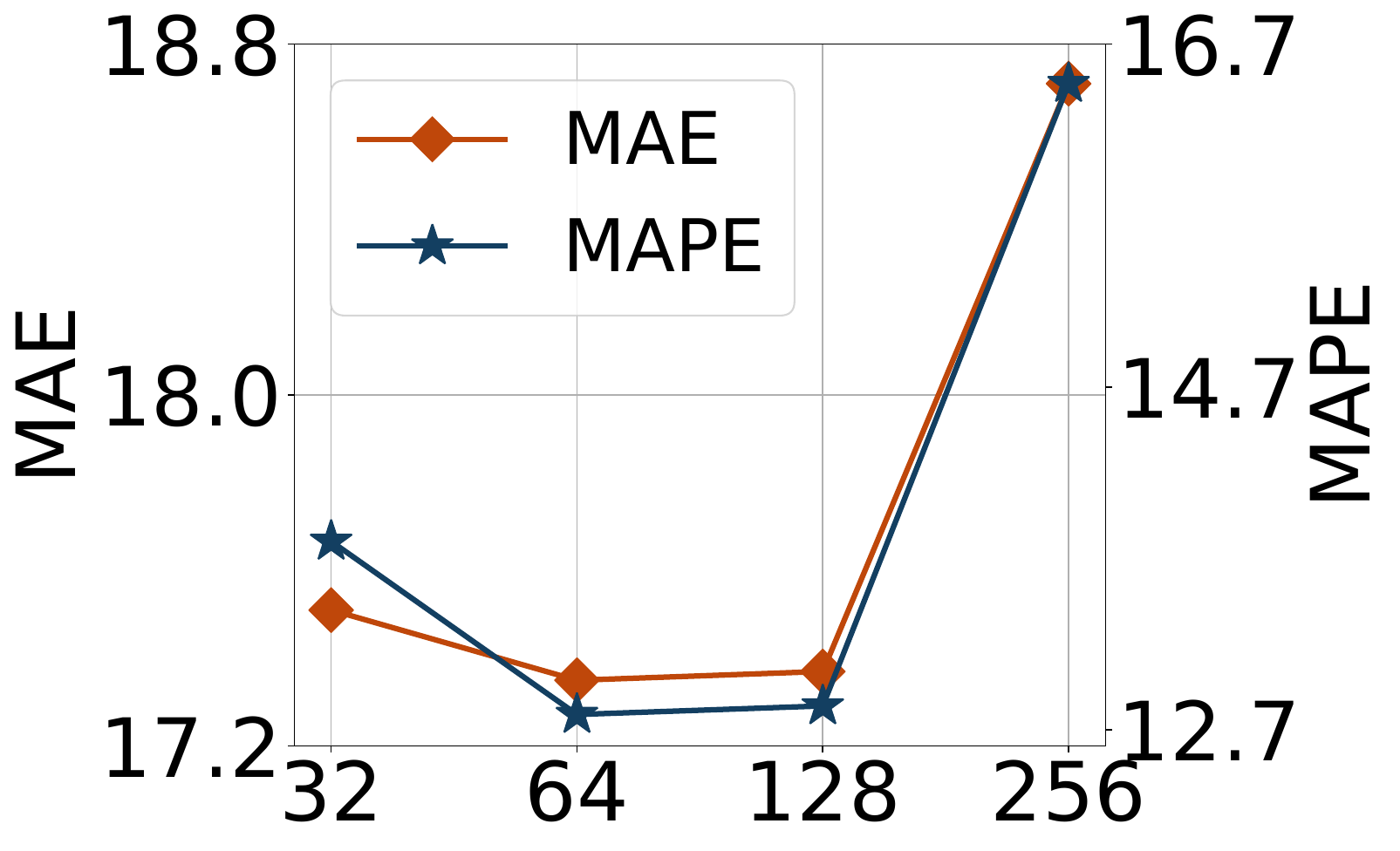}
        \caption{Dimensions on CA}
        \label{exp:cadims}
      \end{subfigure}%
      \vspace{-5pt}
      \caption{Hyper-parameter study.}
      \label{exp:param}
      \vspace{-5pt}
\end{figure}

Figure~\ref{exp:param} draws the impact of hyper-parameters on the representative GBA and CA datasets. We search the number of attention layers and the number of input fully-connected dimensions from a search space of [$1,3,5,7$] and [$32,64,128,256$]. First, the performance of \M~ improves as the layers of the encoder increase and tends to be best when there are $5$ layers. Second, when the number of input dimensions is $64$ and $128$ on GBA and CA datasets, \M~ achieves the best performance. We can observe that the small model is enough to learn spatio-temporal knowledge in large-scale datasets due to rich patterns in big data.

\begin{figure}[t]
    \centering
    \begin{subfigure}{0.45\linewidth}
        \includegraphics[width=\linewidth]{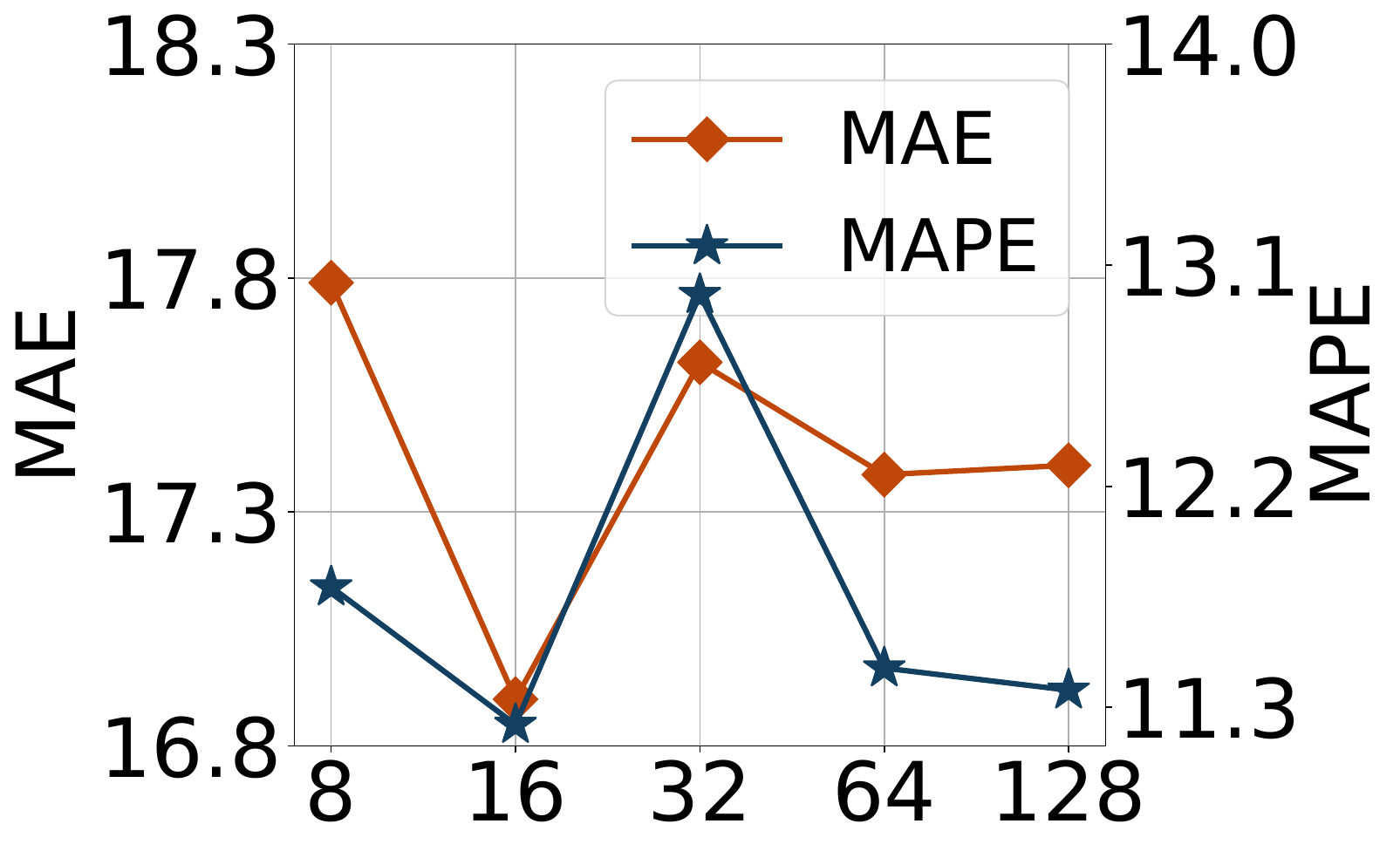}
        \caption{Patches on SD}
        \label{exp:sd}
      \end{subfigure}
      \hspace{2mm}
      \begin{subfigure}{0.45\linewidth}
        \includegraphics[width=\linewidth]{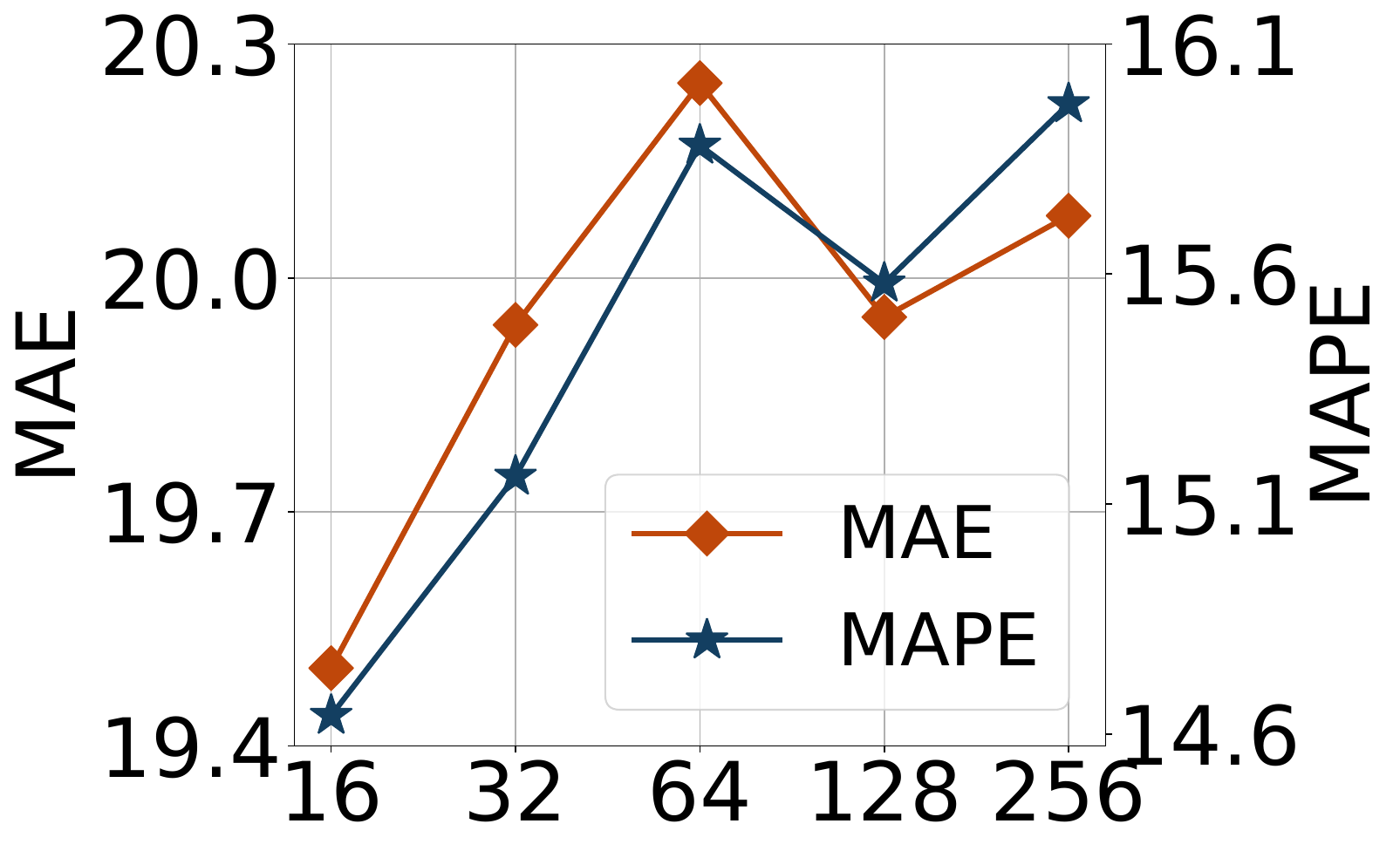}
        \caption{Patches on GBA}
        \label{exp:gba}
      \end{subfigure}%

      \begin{subfigure}{0.45\linewidth}
        \includegraphics[width=\linewidth]{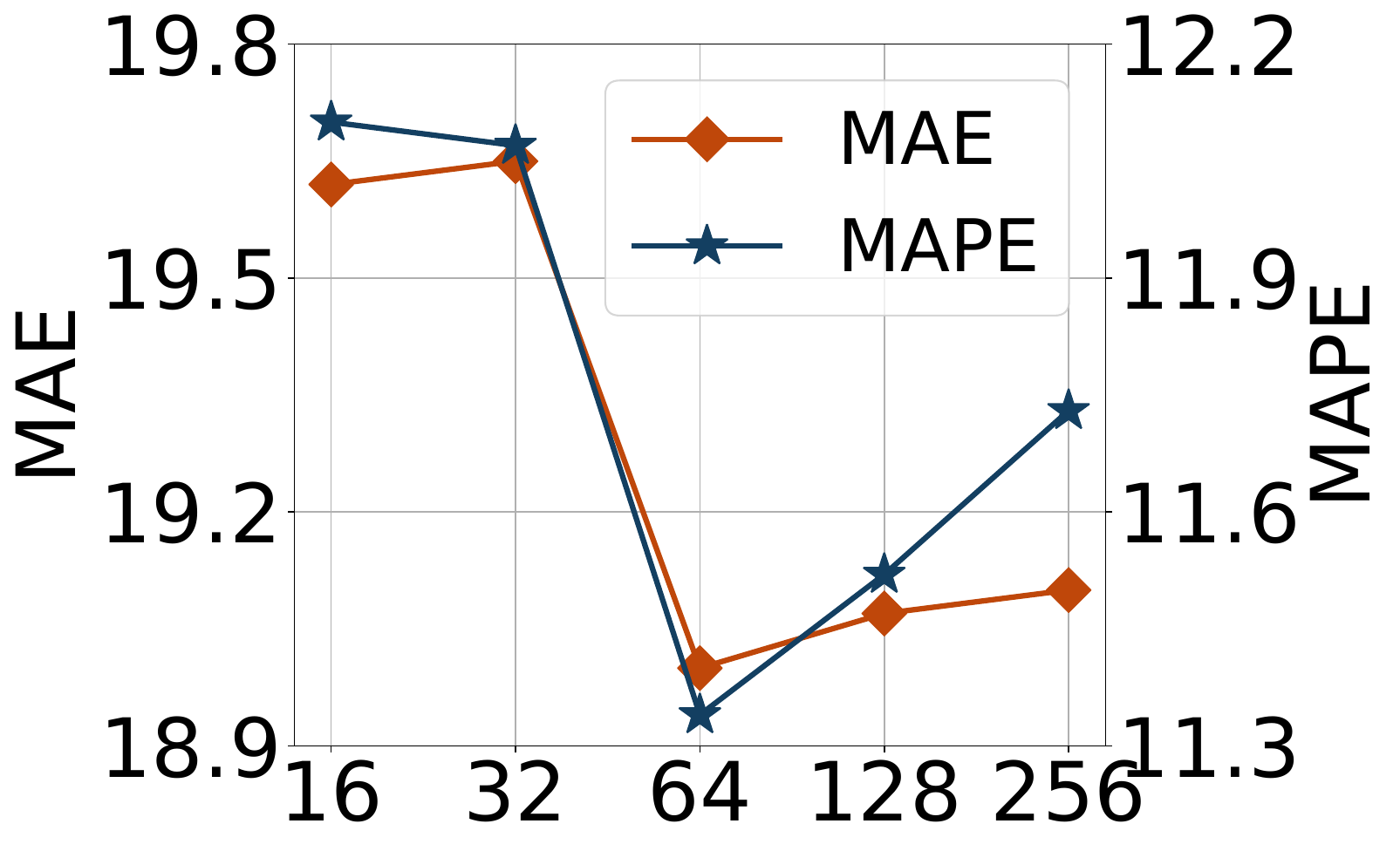}
        \caption{Patches on GLA}
        \label{exp:gla}
      \end{subfigure}
      \hspace{2mm}
      \begin{subfigure}{0.45\linewidth}
        \includegraphics[width=\linewidth]{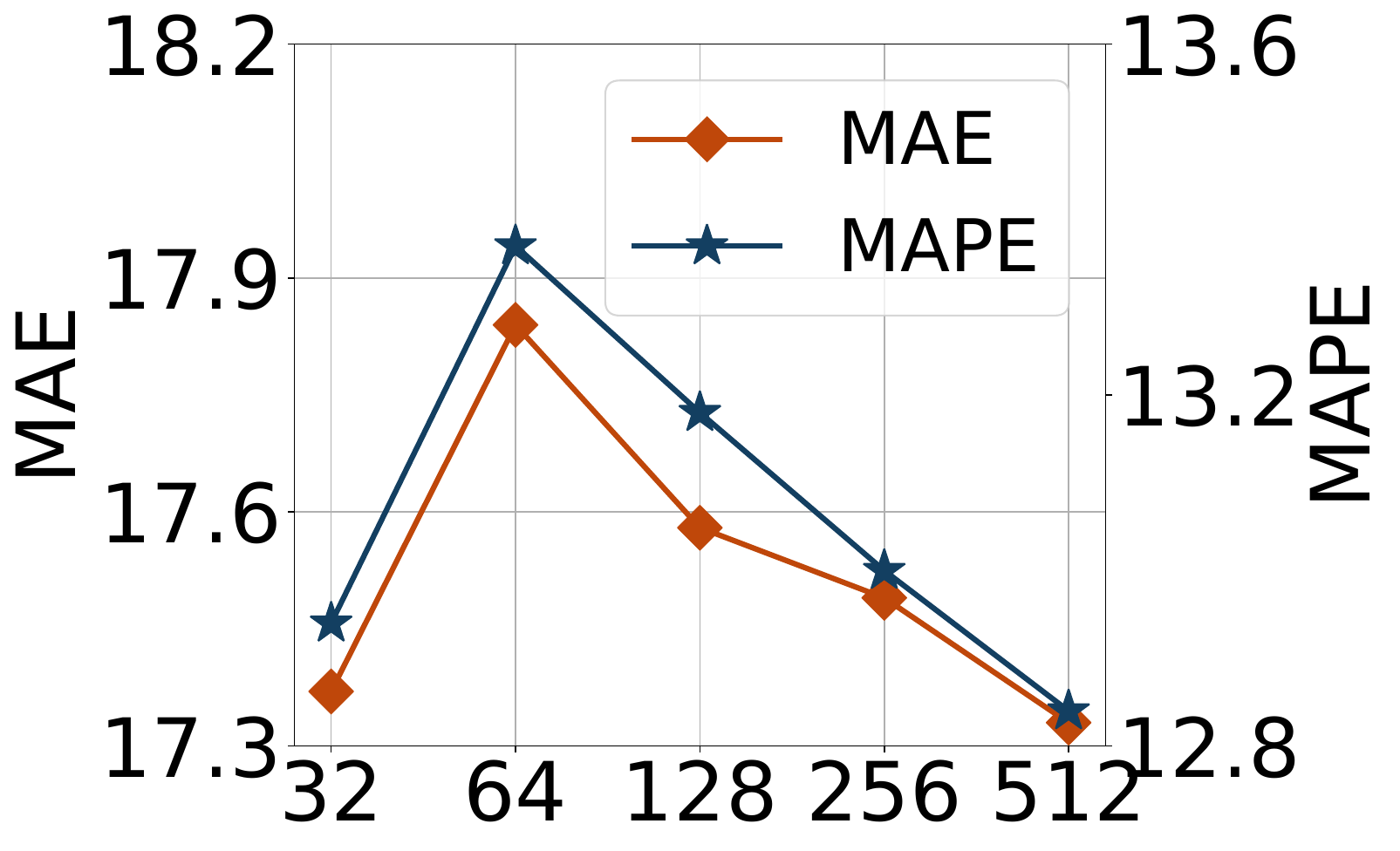}
        \caption{Patches on CA}
        \label{exp:ca}
      \end{subfigure}%
      \vspace{-5pt}
      \caption{The influence of changing the number of patches.}
      \label{exp:patch}
\end{figure}

Moreover, we search the number of patches from a search space of [$8,16,32,64,128,256,512$] as shown in Figure~\ref{exp:patch}. \M~ with $16$, $16$, $64$, and $512$ patches can achieve the best performance on SD, GBA, GLA, and CA datasets, which points out that the number of patches is positively correlated with the size of the dataset.

\section{CONCLUSIONS}
In this paper, we introduce a novel efficient Transformer framework \M~ from the spatial data management perspective for large-scale traffic forecasting. \M~ first utilizes the leaf KDTree to recursively partition the equilibrium number of irregular traffic points into leaf nodes with interpretability. Then \M~ patches leaf nodes with close distance together through backtracking after padding. Finally, \M~ stacks the depth and breadth attention in the encoder interchangeably to efficiently and dynamically capture spatial information from the patched data with fidelity. Experimental results on four large-scale benchmark datasets demonstrate the superior performance of \M~ against $10$ baselines. We will extend \M~ to other spatio-temporal tasks in the future, such as national air quality prediction.

\section*{ACKNOWLEDGMENTS}
This work is partially supported by NSFC (No. 62472068), Shenzhen Municipal Science and Technology R\&D Funding Basic Research Program (JCYJ20210324133607021), and Municipal Government of Quzhou under Grant (2023D044), and Key Laboratory of Data Intelligence and Cognitive Computing, Longhua District, Shenzhen. This work is also supported by the National Natural Science Foundation of China (No. 62402414), the Guangzhou Industrial Information and Intelligent Key Laboratory Project (No. 2024A03J0628), and Guangdong Provincial Key Lab of Integrated Communication, Sensing and Computation for Ubiquitous Internet of Things (No. 2023B1212010007).

\clearpage
\balance
\bibliographystyle{ACM-Reference-Format}
\bibliography{patchstg}

\end{document}